\documentclass[10pt,twocolumn,letterpaper]{article}

\usepackage[accsupp]{axessibility} %

\usepackage{cvpr}              %

\usepackage[dvipsnames]{xcolor}
\newcommand{\Eq}[1]  {Eq.~(\ref{equ:#1})}
\newcommand{\Eqs}[1] {Eqs.~(\ref{equ:#1})}
\newcommand{\Fig}[1] {Fig.~\ref{fig:#1}}
\newcommand{\Figs}[1]{Figs.~\ref{fig:#1}}
\newcommand{\Tbl}[1]  {Table~\ref{tab:#1}}

\newcommand{\Sec}[1] {Sec.~\ref{sec:#1}}

\newcommand{\Etal}{{\textit{et~al.}}}
\newcommand{\Alg}[1] {Algorithm~\ref{alg:#1}}

\DeclareMathOperator*{\argmin}{arg\,min}

\definecolor{cvprblue}{rgb}{0.21,0.49,0.74}
\usepackage[pagebackref,breaklinks,colorlinks,citecolor=cvprblue]{hyperref}
\usepackage{algpseudocode}
\usepackage{algorithm}
\usepackage{algorithmicx}
\usepackage{wrapfig}
\usepackage{lipsum}
\algdef{SE}[DOWHILE]{Do}{doWhile}{\algorithmicdo}[1]{\algorithmicwhile\ #1}

\newcommand\blfootnote[1]{%
  \begingroup
  \renewcommand\thefootnote{}\footnote{#1}%
  \addtocounter{footnote}{-1}%
  \endgroup
}

\def\paperID{14952} %
\def\confName{CVPR}
\def\confYear{2024}

\title{Discontinuity-preserving Normal Integration with Auxiliary Edges}

\author{
Hyomin Kim$^*$
\hspace{1.5cm}
Yucheol Jung$^*$
\hspace{1.5cm}
Seungyong Lee\\
\vspace*{-5pt}\\
POSTECH\\
{\tt\small
\{min00001, ycjung, leesy\}@postech.ac.kr}\\
}

\begin{document}

\twocolumn[{
\renewcommand\twocolumn[1][]{#1}
\maketitle
\centering

\vspace{-20pt}
\includegraphics[width=1.0\textwidth, height=0.50\textwidth]{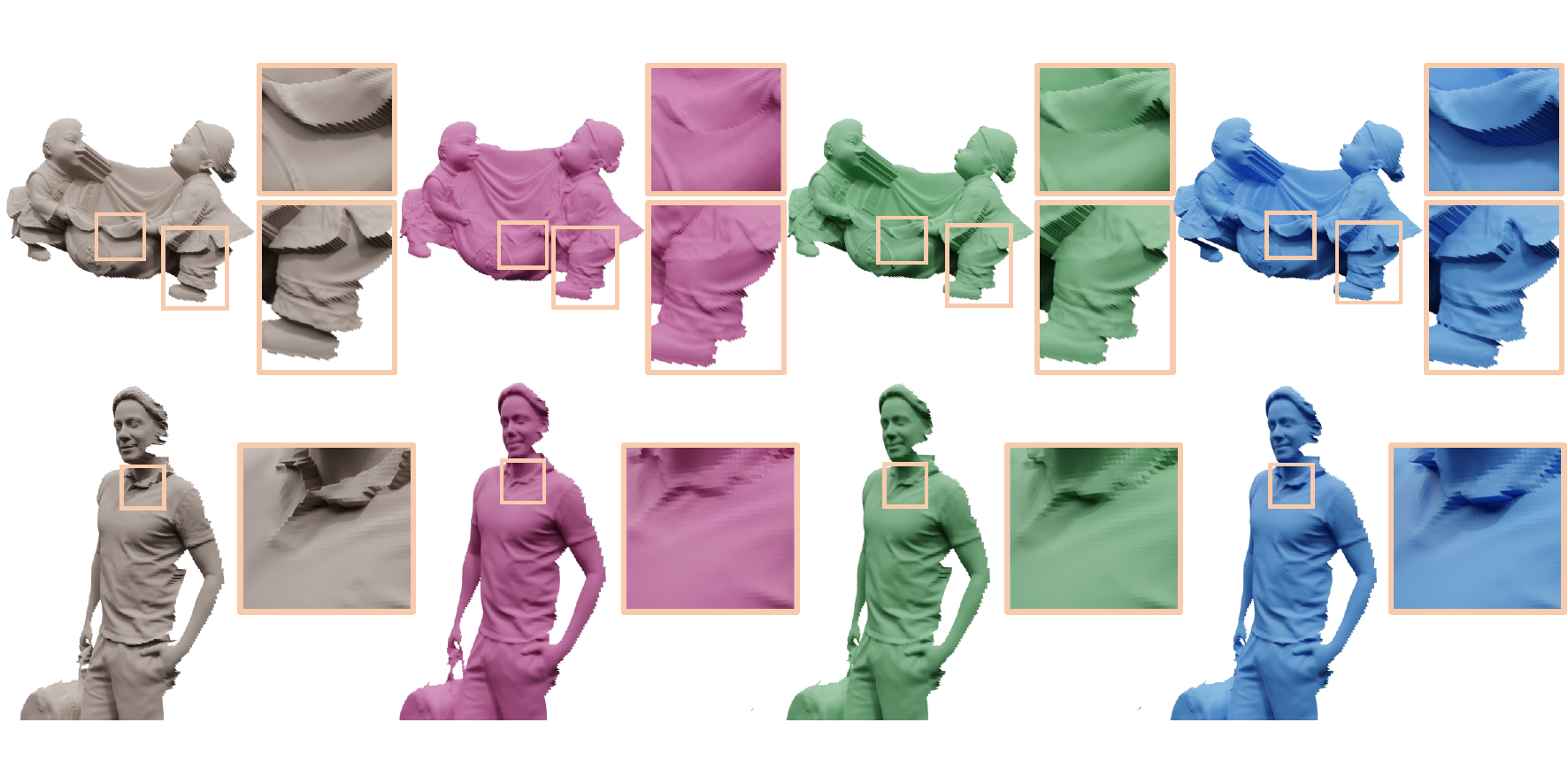} \\
\vspace{-20pt}
\hspace{10pt} Ground truth \hspace{70pt} Poisson \cite{horn1986variational} \hspace{80pt} BiNI \cite{cao2022bilateral} \hspace{85pt} Ours \hspace{50pt}
\vspace{-2pt}
\captionof{figure}{Our normal integration recovers a depth map from an input normal map while preserving surface discontinuities. Compared to the state-of-the-art discontinuity-preserving normal integration method, BiNI \cite{cao2022bilateral}, our method recovers small discontinuities accurately.\vspace{2em}} %
\label{fig:teaser}
}]

\blfootnote{$^*$Equal contribution.}
\begin{abstract}
Many surface reconstruction methods incorporate normal integration, which is a process to obtain a depth map from surface gradients. In this process, the input may represent a surface with discontinuities, e.g., due to self-occlusion. To reconstruct an accurate depth map from the input normal map, hidden surface gradients occurring from the jumps must be handled. To model these jumps correctly, we design a novel discretization scheme for the domain of normal integration. Our key idea is to introduce auxiliary edges, which bridge between piecewise-smooth patches in the domain so that the magnitude of hidden jumps can be explicitly expressed. Using the auxiliary edges, we design a novel algorithm to optimize the discontinuity and the depth map from the input normal map. Our method optimizes discontinuities by using a combination of iterative re-weighted least squares and iterative filtering of the jump magnitudes on auxiliary edges to provide strong sparsity regularization. Compared to previous discontinuity-preserving normal integration methods, which model the magnitudes of jumps only implicitly, our method reconstructs subtle discontinuities accurately thanks to our explicit representation of jumps allowing for strong sparsity regularization.
\end{abstract}

\section{Introduction}

Normal integration is a fundamental problem in 3D shape reconstruction. Most photometric stereo frameworks estimate the surface normal of objects, and normal integration recovers the depth of the object from the estimated normal map \cite{ackermann2015survey}. Normal integration is often used in single-image 3D reconstruction to reconstruct a detailed surface because a high-quality normal map is relatively easy to infer from an image compared to a fully detailed 3D surface \cite{xiu2023econ}.

A challenge in normal integration is the fact that the normal map may represent a depth map with discontinuities. Discontinuities in a depth map are universally observed in captures of real-world objects due to self-occlusions and the hidden portions of surfaces. 
Not handling the discontinuity and assuming the surface is continuous everywhere introduces global distortions in normal integration results. In this work, we design a discontinuity-preserving normal integration method to automatically detect and handle surface discontinuity. 

Previous discontinuity-preserving normal integration methods model such jumps by locating and weakening, or removing the effect of edges located across the discontinuity in the image grid. In this work, we design a new approach that models the jumps at discontinuity by \textit{adding} edges corresponding to the jumps, instead of \textit{removing} edges from the original image grid. We implement this design by introducing {\em auxiliary} edges that connect neighboring pixels in the original grid.

Our intuition behind this additive design is that the surface discontinuities can be viewed as additional hidden gradients that are not addressed by the input normal map. Unlike previous approaches \cite{leonardis_what_2006, tai-pang_wu_visible_2006, cao2022bilateral} that models the jumps implicitly without considering what the values of the jumps across discontinuities should be, our approach enables explicit control over the locations and the degrees of the jumps. Benefiting from this explicit control for the jumps, we design an iterative optimization scheme that recovers sparse surface discontinuity from the input normal map by combining iterative re-weighted least squares (IRLS) \cite{chartrand2008iteratively} and iterative filtering of the jump magnitudes on the auxiliary edges. Compared to the state-of-the-art method \cite{cao2022bilateral}, our approach accurately detects challenging subtle discontinuities thanks to our sparsity regularization (\Fig{teaser}).

In summary, our technical contributions are as follows:
\begin{itemize}
    \item We propose a novel normal integration framework based on a discrete graph with auxiliary edges that explicitly models the locations and magnitudes of jumps across surface discontinuities. %
    \item We design a novel sparsity-regularized iterative optimization method for discontinuity and the depth map by combining IRLS and iterative filtering of jump magnitudes.
\end{itemize}

\section{Problem Statement}

Given a single-view normal map $\mathbf{n}$ of an object, normal integration is a process to recover the depth map $\mathbf{d}$ from $\mathbf{n}$.
First, the depth gradient $\hat{\mathbf{g}}$ is calculated from the normal map $\mathbf{n}$. Assuming normal maps captured using orthographic cameras, the gradient map is
\begin{equation}
    \hat{\mathbf{g}} = \dfrac{\mathbf{n}_{xy}}{n_z}
    \label{equ:main-pde}
\end{equation}
with the $xy$ component $\mathbf{n}_{xy} = (n_x, n_y)$ and the $z$ component $n_z$. Then, the depth map $\mathbf{d}$ is reconstructed using the relation
\begin{equation}
    \nabla \mathbf{d} = \hat{\mathbf{g}},
    \label{equ:conversion}
\end{equation} 
where $\nabla\mathbf{d}$ is the gradients of $\mathbf{d}$. In the case of using a perspective camera, a similar relation can be formed \cite{cao2022bilateral}. The perspective case only requires two changes: 1) $n_z$ is scaled according to the image coordinates and 2) $\log\mathbf{d}$ is used instead of $\mathbf{d}$ in the optimization; then the final depth is recovered by applying exponential to the log values. The details for the perspective case can be found in the supplementary document. In this paper, for notational convenience, we describe our algorithm using $n_z$ and $\mathbf{d}$.

Ideally, if the vector field $\hat{\mathbf{g}}$ is the true gradient map of the depth map to be reconstructed, the depth value at a position can be obtained by accumulating $\hat{\mathbf{g}}$ along any continuous path to the position from a position with a known depth \cite{queau_normal_2018}. However, for real-world data, $\hat{\mathbf{g}}$ is rarely a true gradient field for the depth map, causing different depths for different integration paths. This error can be caused by multiple factors \cite{saracchini2012robust}, including sensor noise, quantization error, and surface discontinuity. 
As a result, variational methods have been proposed \cite{queau_variational_2018} to implicitly edit the input gradient $\hat{\mathbf{g}}$ using an editing $\mathbf{g}'$ so that the vector field becomes a true depth gradient field $\mathbf{g}$;
\begin{equation}
\mathbf{g} = \hat{\mathbf{g}} + \mathbf{g}'.
\label{equ:decomposition}
\end{equation}

Multiple design choices can be considered to build the gradient editing $\mathbf{g}'$. Among them, modeling surface discontinuity is a critical one that should be considered to avoid large errors in reconstructions of real-world objects. For example, when the target surface contains large jumps along the view direction, e.g., due to self-occlusions, the jump cannot be effectively described using only the surface normal captured in an image. As a result, an integration path on the image grid crossing the occlusion boundary and another path not crossing the boundary produce largely different results. If the editing $\mathbf{g}'$ correctly adds the jump at the discontinuous boundary, the two integration paths will produce the same results. In this work, given the input normal map $\mathbf{n}$, we aim to optimize $\mathbf{g}'$ to obtain a correct depth map as a result of normal integration.

In the next section, we review previous normal integration frameworks and their implications in terms of gradient editing and handling of surface discontinuities.

\section{Related Work}

\subsection{Normal integration}

Normal integration is an extensively studied topic in computer vision. Refer to \cite{queau_normal_2018} and \cite{queau_variational_2018} for a survey of methods that aim to achieve various desirable properties for normal integration, such as fast computation or robustness to noisy input. In this work, we focus on handling discontinuity on the surface. We first review a few notable methods for normal integration in general, then review various previous approaches for discontinuity-preserving normal integration.

A standard approach to reconstruct the depth satisfying 
\Eq{main-pde}, despite $\hat{\mathbf{g}}$ not being a true gradient map, 
is to find a least squares solution minimizing the difference between the gradient of the optimized depth and $\hat{\mathbf{g}}$. This approach of Horn and Brooks \cite{horn1986variational} is equivalent to solving a Poisson equation. As a result, the depth obtained from this approach implicitly applies minimal editing to the input gradient map by minimizing the $l_2$ norm of $\mathbf{g}'$. However, the minimization of $l_2$ norm produces dense $\mathbf{g}'$, whereas discontinuities should be located sparsely at occlusion boundaries. Consequently, this approach cannot correctly handle the discontinuities.

Xie \Etal~\cite{xie_surface--gradients_2014} proposed a normal integration method based on discrete geometry processing (DGP) where a quadrilateral is constructed for each pixel in the normal map, and normal integration is performed using the constructed mesh. Yet, this DGP method does not handle surface discontinuities. Our normal integration method constructs a similar discrete grid using a quadrilateral per pixel. In contrast to the previous DGP method, we introduce auxiliary edges that bridge between pixels to model discontinuities on the surface.

\subsection{Discontinuity-preserving normal integration}

Surface discontinuity in normal integration has been handled in various ways. An early approach \cite{karacali_reconstructing_2003} proposed 
automatic localization of discontinuities using the error residuals between the input gradients and the gradients of the surface reconstructed assuming no discontinuity. Agrawal and Raskar \cite{agrawal_algebraic_2005} considered the integration domain as a graph and cut the graph based on the curl of the input gradients. Fraile \cite{fraile_combinatorial_2006} proposed construction of an integration tree by computing a minimum spanning tree of a weighted graph with weights computed from the gradient map, e.g., using the principal curvature direction. However, such approaches based on a single analysis of the error residuals or the input gradients may not localize the discontinuity effectively; often, the residuals or the curls highlight only a part of the discontinuity.

To overcome the limitations of single-analysis approaches, iterative optimization methods have been employed for discontinuity-preserving normal integration, where the depth and the location of discontinuities are optimized alternatingly. Agrawal~\Etal~\cite{leonardis_what_2006} proposed a framework to build a spanning tree for integration by iteratively cutting the graph based on the residuals computed from intermediate depth values. Wu~\Etal~\cite{tai-pang_wu_visible_2006} proposed an expectation-maximization algorithm alternating between the discontinuity map and the parameters used to estimate the discontinuity map. 
Quéau~\Etal~\cite{queau_variational_2018} proposed several approaches to optimize sparse discontinuity using $l_1$ gradient minimization, sparse gradient minimization using robust estimators, integration by anisotropic diffusion, and the minimization of Mumford-Shah functional.

Recently, Cao~\Etal~\cite{cao2022bilateral} proposed a bilaterally weighted functional to perform discontinuity-preserving normal integration and demonstrated a large improvement over the methods of Quéau~\Etal~\cite{queau_variational_2018} in terms of depth reconstruction error. The bilateral weights are iteratively computed based on intermediate depth reconstruction results so that the gradients from the pairs of depth pixels crossing the occlusion boundary weakly influence the normal integration. However, as the bilateral weights provide only a limited level of sparsity using smooth Heaviside function, small discontinuities may not be detected.

Our method is also based on the alternating optimization of the depth and discontinuity. Unlike previous approaches handling the gradient at discontinuity only implicitly via graph cuts or weight computation for functionals, we explicitly model the gradients across the surface discontinuity using auxiliary edges. Our approach detects small discontinuities accurately via sparsity regularization based on the explicit filtering of gradients at auxiliary edges, allowing more accurate reconstruction of the depth map.

\section{Normal Integration with Auxiliary Edges}

Given the gradient field $\hat{\mathbf{g}}$ in \Eq{main-pde} calculated from the input normal map $\mathbf{n}$, our goal is to optimize the gradient editing $\mathbf{g}'$ in \Eq{decomposition} to create correct jumps across surface discontinuities when the gradient field is integrated to obtain the depth map $\mathbf{d}$. As the jumps for the discontinuities would occur only at the occlusion boundary, we assume $\mathbf{g}'$ to be zero almost everywhere. In other words, the $l_0$ norm of $\mathbf{g}'$ should be minimized while allowing the resulting depth $\mathbf{d}$ to match the gradient field $\hat{\mathbf{g}}$ as much as possible:
\begin{equation}
    \mathbf{g}', \mathbf{d} = \argmin_{\mathbf{g}', \mathbf{d}} E_{data}(\hat{\mathbf{g}}+\mathbf{g}', \mathbf{d}) + \lVert \mathbf{\mathbf{g}'} \rVert_0,
    \label{equ:master}
\end{equation} where $E_{data}$ refers to the \textit{data} term that measures the discrepancy between the gradient map $\hat{\mathbf{g}} + \mathbf{g}'$ and the normal integration result $\mathbf{d}$. The data term requires $\mathbf{g}'$ not to be zero everywhere: the jumps for the discontinuity must be properly added to minimize the discrepancy.

$l_0$-regularized optimization using strategies based on auxiliary variables has been studied in a wide range of applications, including image smoothing \cite{xu2011image} and 3D shape filtering \cite{he2013mesh}. The minimization of $l_0$ norm is NP-hard, and different formulations to obtain a local optimum have been proposed depending on the problem domain. In this work, we design a novel $l_0$-regularized optimization scheme specialized for normal integration using a discrete grid with auxiliary edges.

\begin{figure}
    \centering
    \includegraphics[width=\linewidth]{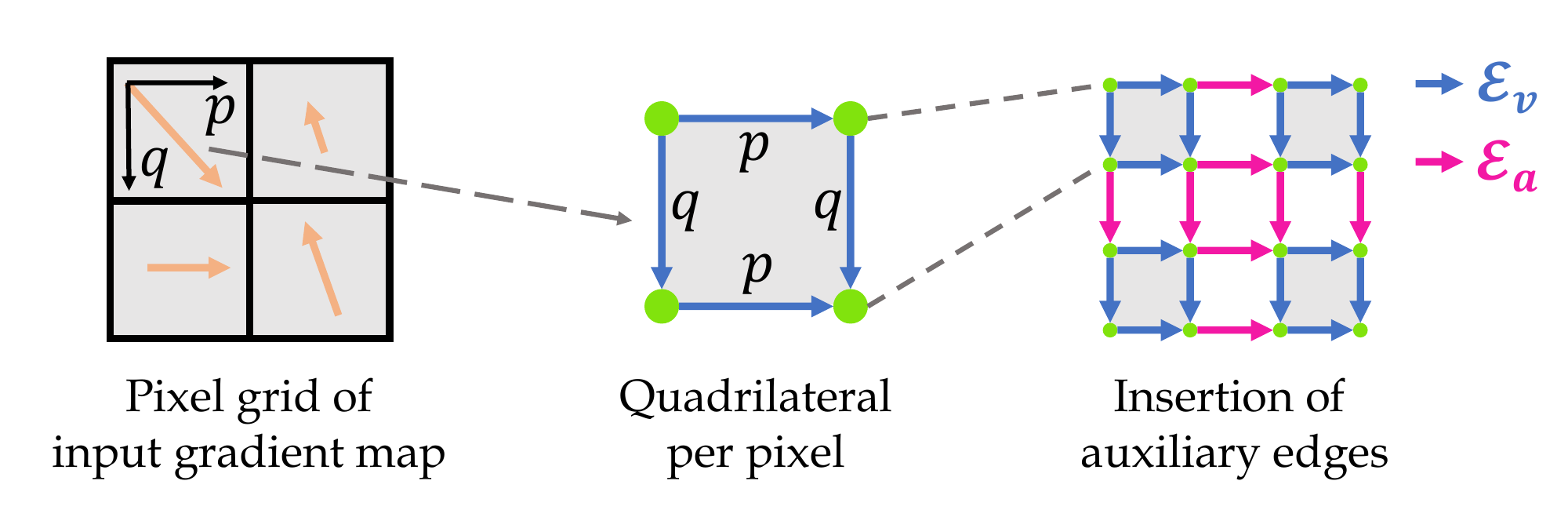}
    \caption{Construction of our graph for normal integration. For each pixel in the input gradient map, we construct a quadrilateral with four vertices for depth and their edges $\mathcal{E}_v$ for directional derivatives. The quadrilaterals are bridged with auxiliary edges $\mathcal{E}_a$ that model gradients across the discontinuity.}
    \label{fig:discrete-grid}
\end{figure}

\begin{figure}
    \centering
    \includegraphics[width=0.7\linewidth]{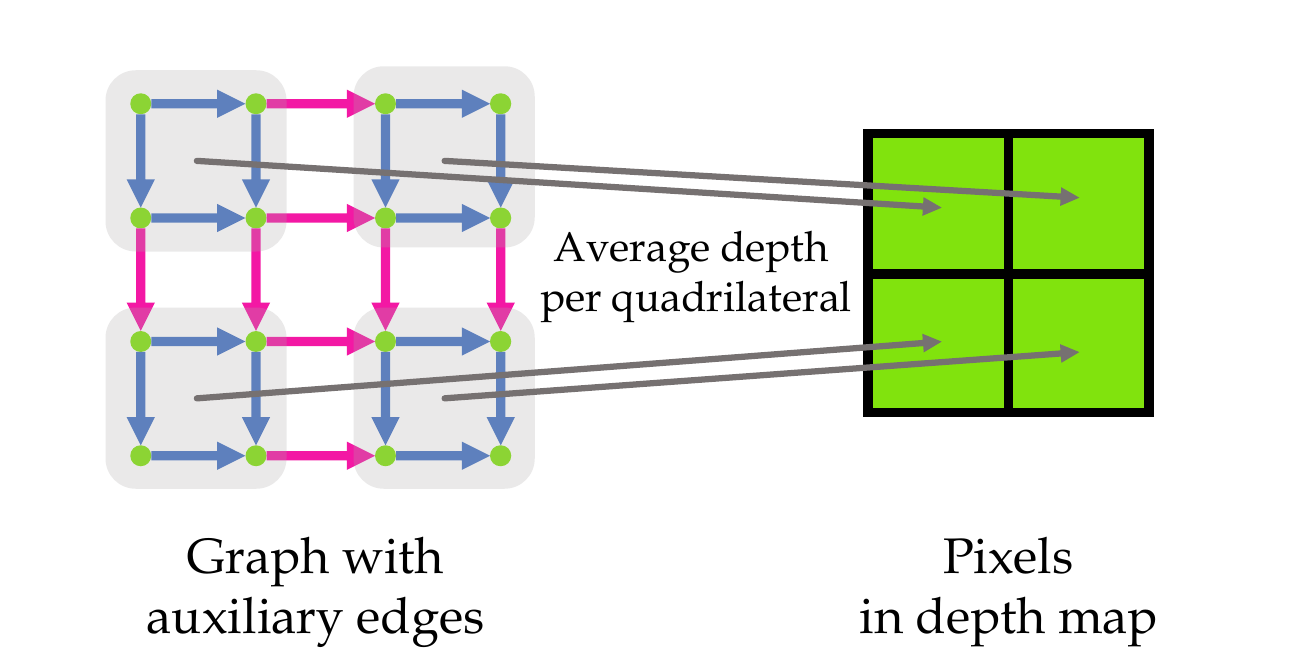}
    \caption{Construction of depth map from the normal integration results on the graph. The depth values on the four vertices of each quadrilateral are averaged and assigned to a depth pixel.}
    \label{fig:depth-map}
\end{figure}

Our key idea is to model the gradients $\hat{\mathbf{g}}$ and the editing $\mathbf{g}'$ on separate finite elements in a directional graph that defines the discrete domain for normal integration. In this graph, each vertex is assigned a depth, and an edge is assigned a directional derivative. 
We first construct a quadrilateral for each pixel in the input normal map. Then, we bridge each quadrilateral with auxiliary edges (\Fig{discrete-grid}). The set of edges $\mathcal{E}_v$ belonging to the quadrilaterals are assigned the projection of gradient $\hat{\mathbf{g}}$ to the edges, and the set of auxiliary edges $\mathcal{E}_a$ are assigned the directional gradients $\mathbf{g}'$ across quadrilaterals. After performing the normal integration on this graph, we construct the depth map on a pixel grid by averaging the depth values of vertices for each quadrilateral (\Fig{depth-map}).

On this graph with auxiliary edges, the data term $E_{data}$ measuring the discrepancy between the gradient map and the depth can be defined as a sum of two terms, assuming an orthographic camera:
\begin{equation}
    E_{data}(\hat{\mathbf{g}}+\mathbf{g}', \mathbf{d}) = E_v(\hat{\mathbf{g}}, \mathbf{d}) + E_a(\mathbf{g}', \mathbf{d})
    \label{equ:energy-data}
\end{equation} with
\begin{equation}
    E_v(\hat{\mathbf{g}}, \mathbf{d}) = \sum_{e \in \mathcal{E}_v} \lVert n_z(e)(D_e \mathbf{d}  - \hat{\mathbf{g}}(e)) \rVert^2,
    \label{equ:ev}
\end{equation}
\begin{equation}
    E_a(\mathbf{g}', \mathbf{d}) = \sum_{e \in \mathcal{E}_a} \lVert n_z'(e)(D_e \mathbf{d}  - \mathbf{g}'(e)) \rVert^2,
    \label{equ:ea}
\end{equation} where $n_z(e)$ is the $z$ component of the normal in the pixel containing the edge $e$, $D_e \mathbf{d}$ is the directional derivative of $\mathbf{d}$ at $e$ along the edge direction, and $\hat{\mathbf{g}}(e)$, $\mathbf{g}'(e)$ are the gradient values at edge $e$. The residuals for $E_v$ are scaled with $n_z$ for numerical stability; occasionally $n_z \approx 0$ and then $\hat{\mathbf{g}}$ obtained from \Eq{main-pde} may diverge to infinity. We scale the residuals for $E_a$ using a constant $n_z'=1$ as the 3D surface normals corresponding to auxiliary edges are not clearly defined.
Formulations of \Eqs{ev} and (\ref{equ:ea}) in the perspective case can be found in the supplementary document.

We model surface discontinuities as sparse gradient editing $\mathbf{g}'$, which are contained in the auxiliary edges. We incorporate the sparsity constraint of \Eq{master} by modifying the data term $E_a$ for auxiliary edges. The modified energy is
\begin{equation}
    E_{disc}(\mathbf{g}', \mathbf{d}, \mathbf{w}) = \sum_{e \in \mathcal{E}_a} w_{e}\lVert n_z'(e)(D_e \mathbf{d}  - \mathbf{g}'(e))\rVert^2.
    \label{equ:edisc}
\end{equation} 
As directly minimizing $\lVert\mathbf{g}'\rVert_0$ is hard, we instead iteratively build sparse gradients $\mathbf{g}'$ based on depth $\mathbf{d}$ obtained with sparsity-regularized optimization. The variable weight $w_e$ enables sparsity regularization for the residuals in $E_{disc}$ using iteratively reweighted least squares (IRLS) \cite{chartrand2008iteratively}, which is often employed to obtain sparse solutions that minimize $l_p$-norm for $0 < p < 1$.
With IRLS, optimization of $\mathbf{d}$ could produce non-zero residuals in $E_{disc}$ at sparse discontinuities, where $\mathbf{d}$ is allowed to be less affected by the values of $\mathbf{g}'$ that may not represent
the desired sizes of jumps yet.
We then update $\mathbf{g}'$ with explicitly filtered gradients of $\mathbf{d}$ to modulate jump sizes and introduce additional discontinuities.
The iterative update is described in the next section.

\section{Discontinuity Optimization via Iterative Reweighting and Gradient Filtering}

Our iterative optimization involves three variables: the depth map $\mathbf{d}$, the weights $\mathbf{w}$, and the gradient editing $\mathbf{g}'$ on the auxiliary edges. The variables are updated in an alternating fashion using \Alg{iterative}.

\begin{algorithm}
\caption{Update of $\mathbf{d}$, $\mathbf{w}$, and $\mathbf{g}'$}\label{alg:iterative}
\begin{algorithmic}
\State $n \gets 0$
\State $\mathbf{g}' \gets \mathbf{0}$
\State $\mathbf{w} \gets \mathbf{1}$
\State $\lambda_c = 0.5(\lambda_{soft} + \lambda_{hard})$
\Do
    \For{$\lambda \in [\lambda_{soft}, \lambda_c, \lambda_{hard}, \lambda_c]$}
    \State $\mathbf{d} \gets \textbf{argmin } E_v(\hat{\mathbf{g}}, \mathbf{d}) + \lambda E_{disc}(\mathbf{g}', \mathbf{d}, \mathbf{w})$
    \State $\mathbf{w} \gets \textbf{ reweight}(\mathbf{d})$
    \State $\mathbf{g}' \gets \textbf{filter}(\mathbf{d})$
    \State $n \gets n+1$
    \EndFor
\doWhile{$n < N_{max}$}
\end{algorithmic}
\end{algorithm}

\begin{figure*}
    \centering
    \includegraphics[width=\linewidth]{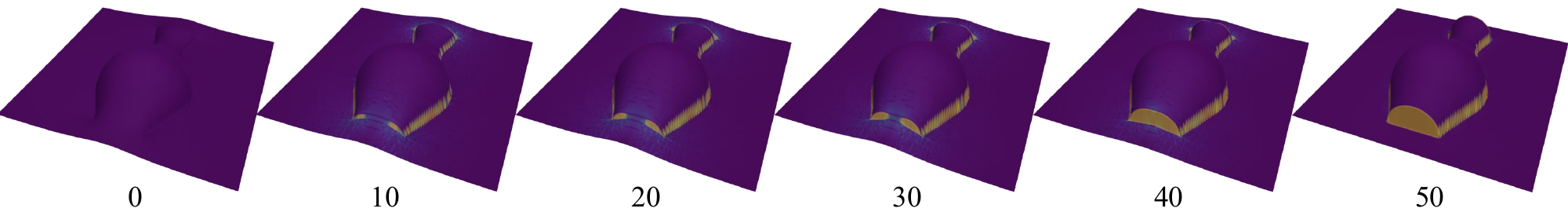}
    \caption{Visualization of $\mathbf{g}'$ and $\mathbf{d}$ across iterative optimization. The gradients $\mathbf{g}'$ on auxiliary edges are visualized as color; bright color means higher magnitude. The depth $\mathbf{d}$ is shown as the height on the surface. The numbers at the bottom show the number of iterations $n$.}
    \label{fig:iterative}
\end{figure*}

\paragraph{Solving for depth}
The minimization of $E_v + \lambda E_{disc}$ can be formulated as a weighted least squares (WLS) problem, which forms a linear equation of the form $\mathbf{A}^{T}\mathbf{W}\mathbf{A}\mathbf{d} = \mathbf{A}^T\mathbf{b}$ such that $\mathbf{A}\mathbf{d} - \mathbf{b}$ represents an array of residuals from all edges. The linear equation is solved using a conjugate gradient solver \cite{hestenes1952methods}.

Initially, we start with $\mathbf{g}' = \mathbf{0}$ and $\mathbf{w} = \mathbf{1}$ to solve for $\mathbf{d}$. In this case, we are simply solving an ordinary least squares problem as in the Poisson-equation-based method \cite{horn1986variational}. Therefore, the initial depth would prefer dense and wide-spread residuals of $E_{disc}$ so that the $l_2$ norm of $D_e \mathbf{d}$ is minimized, which means most pixels would have a small discontinuity nearby.

Given an intermediate solution $\mathbf{d}$, the \textbf{reweight} and \textbf{filter} steps are performed to update the weights $\mathbf{w}$ and the gradients $\mathbf{g}'$ for obtaining a more desirable solution in the next iteration. The updates of $\mathbf{g}'$ and $\mathbf{d}$ through this iterative process are visualized in \Fig{iterative}.

\paragraph{Reweighting}
Using the depth map $\mathbf{d}$, we calculate the weights for the next step using directional derivatives $D_e$;
\begin{equation}
    w_{e} = \min \left[ \dfrac{1}{(D_e \mathbf{d})^2}, 1 \right].
\end{equation}
Based on the values $D_e \mathbf{d}$, the reweighting assigns higher weights for small gradients and lower weights for high gradients. In our case, most auxiliary edges have low jumps with small $D_e \mathbf{d}$ and are assigned high weights so that the small jumps can be suppressed by lowering the differences of $D_e \mathbf{d}$ from $\mathbf{g}'$, which should be zero almost everywhere. On the other hand, edges with 
large $D_e \mathbf{d}$ are allowed to produce even bigger jumps in the next iteration because of the lowered weights.

\paragraph{Gradient filtering}
However, applying the update to $\mathbf{w}$ alone does not guarantee strong enough sparsity for the gradient of $\mathbf{d}$ to properly express surface discontinuities (\Sec{ablation}).
We additionally perform filtering of the gradients $D_e \mathbf{d}$ in the update of $\mathbf{g}'$ to build sparse gradients. 
This direct manipulation of the gradients of auxiliary edges enables more aggressive control of sparsity on the discontinuities in the depth map.

We build a new sparse gradient map $\mathbf{g}'$ by filtering $D_e \mathbf{d}$ with a filter function $f$ defined on auxiliary edges:
\begin{equation}
\mathbf{g}'(e) = f(e) \cdot D_e \mathbf{d},
\end{equation}
where $e \in \mathcal{E}_a$.
We design $f(e)$ to produce a more sparse version of $D_e \mathbf{d}$, with non-zero values located at discontinuities. First, we expect the direction crossing the discontinuity to introduce abrupt change in the tangentness to the view direction. We measure this quality using a function $s(e)$ that calculates the degree of local change in tangentness:
\begin{equation}
    s(e) = [n_z(e_v') - n_z(e_v'')]^2 + \tau,
\end{equation} 
\begin{wrapfigure}[3]{r}{0.42\linewidth}
\vspace{-10pt}
\hspace{-22pt}
\includegraphics[width=1.2\linewidth]{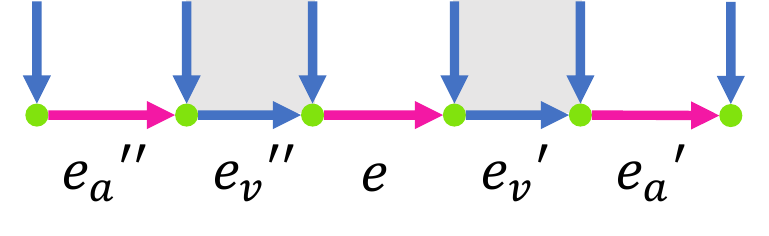}
\end{wrapfigure} 
for a small value $\tau$ and non-auxiliary neighbor edges $e_v'$, $e_v''$ of $e$ along the direction $e$. $n_z$ is used to capture the tangentness with the view direction; a small value means the surface is more tangent to the view direction. Using this scaling function, we calculate scaled gradients $G(e)$:
\begin{equation}
    G(e) = s(e) \cdot D_e \mathbf{d}.
\end{equation} 
Second, in addition to considering the tangentness to the view direction, we also suppress gradients with non-maximal magnitudes; the jump magnitudes for the discontinuities are expected to be locally maximum. Therefore, we calculate the local-maximumness of the squared magnitude $[G(e)]^2$ using the Laplacian $L(e)$:
\begin{equation}
    L(e) = 2[G(e)]^2 - [G(e_a')]^2 - [G(e_a'')]^2
\end{equation} for two neighboring auxiliary edges $e_a'$ and $e_a''$ along the direction of $e$.

\begin{figure*}
    \centering
    \includegraphics[width=1.0\textwidth]{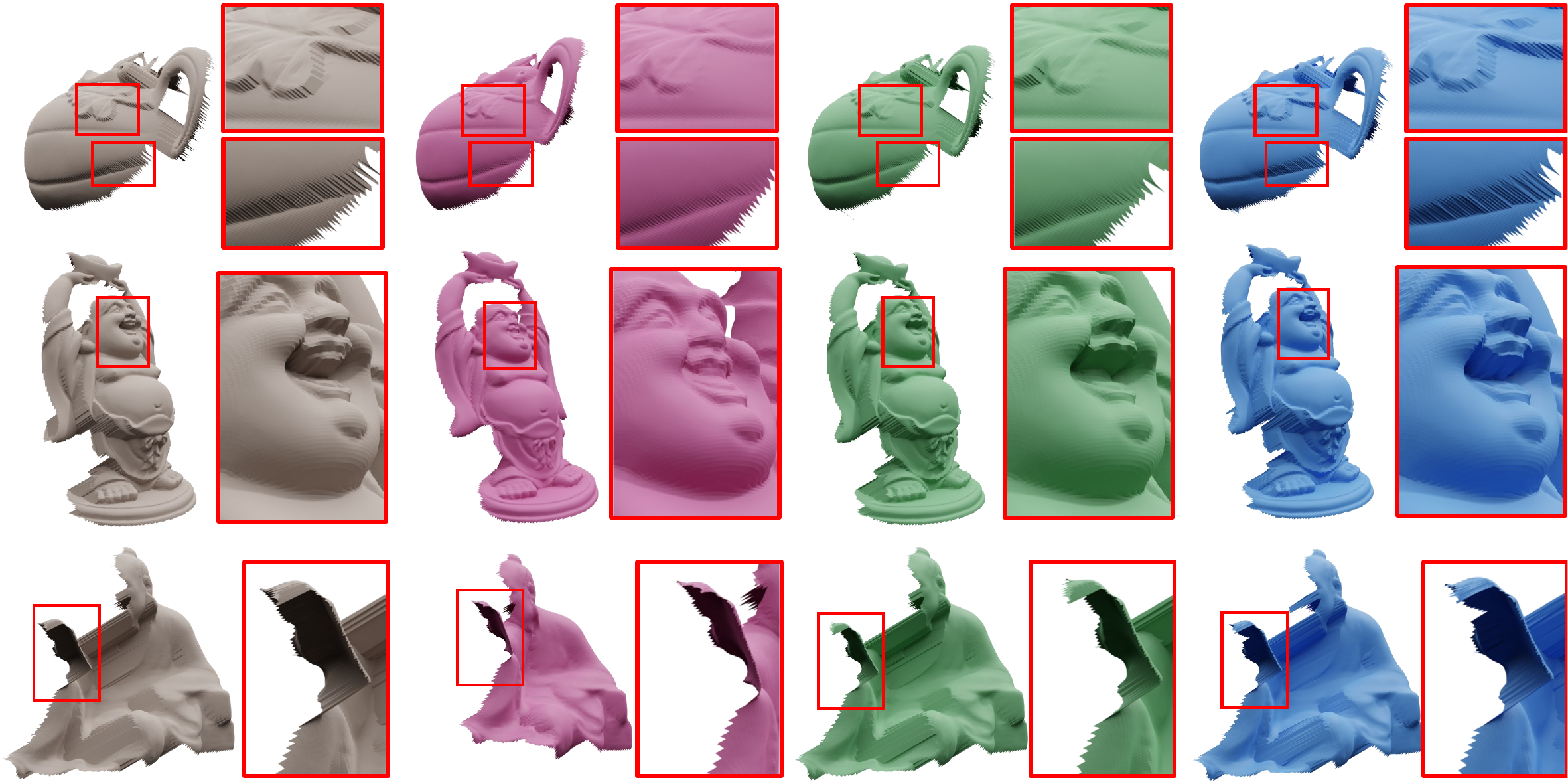} \\
    \hspace{20pt} Ground truth \hspace{70pt} Poisson \cite{horn1986variational} \hspace{80pt} BiNI \cite{cao2022bilateral} \hspace{85pt} Ours \hspace{40pt}
    \caption{Qualitative comparison with previous normal integration methods using the DiLiGeNT dataset. Our method can reconstruct challenging and small discontinuities as well as large discontinuities that are handled by BiNI \cite{cao2022bilateral}.}
    \label{fig:qualitative-diligent}
    \vspace{-6pt}
\end{figure*}

Finally, the filter function $f(e)$ is designed as the thresholding on the local-maximumness to enforce sparsity:
\begin{equation}
    f(e) = 
    \begin{cases}
    1/(1+\textit{exp}(-kL(e)) & L(e)>0 \\
    0 & L(e)\le0
    \end{cases}
\end{equation} for a large value $k$.
This filtering suppresses most gradients from $D_e \mathbf{d}$ except sparse jumps across discontinuities.

\paragraph{Two-pass update with $\lambda_{soft}$ and $\lambda_{hard}$}
Aggressive control of sparsity in $\mathbf{g}'$ via direct gradient manipulation may result in incorrect discontinuities: a high value of $\mathbf{g}'$ may occur inside a continuous surface patch. Still, such erroneous large jumps would be healed automatically via the next depth solve step that produces a depth $\mathbf{d}$ globally consistent with $E_v$, penalizing the discrepancy from the gradients $\hat{\mathbf{g}}$. On the other hand, the depth solve step may also remove valid non-zero values in $\mathbf{g}'$ if the non-zero jump values are not spread wide enough to form a cut that allows a coherent large jump in depth values across the occlusion boundary.

To heal the errors while keeping the valid newborn discontinuities from vanishing, we design a two-pass update scheme, which is inspired by cosine annealing \cite{loshchilov2016sgdr} that is known to facilitate the escape from local optima. 
We alternate between small weight $\lambda_{soft}$ and large weight $\lambda_{hard}$ for $E_{disc}$. In the first pass, we run the update process with $\lambda_{soft}$ to heal any defective gradients in $\mathbf{g}'$.
Then, the second pass reinforces the discontinuities obtained from the first step by updating with $\lambda_{hard}$; optimizing with a strong weight $\lambda_{hard}$ widens the discontinuities in $\mathbf{g}'$, as strong influences from non-zero gradients in $\mathbf{g}'$ promote nearby auxiliary edges to form discontinuities accordingly. Each pass is followed by another pass using the mean of $\lambda_{soft}$ and $\lambda_{hard}$ similarly to cosine annealing. This reinforcement step helps the next iteration with $\lambda_{soft}$ to retain valid discontinuities.

\begin{figure*}
    \centering
    \includegraphics[width=1.00\textwidth]{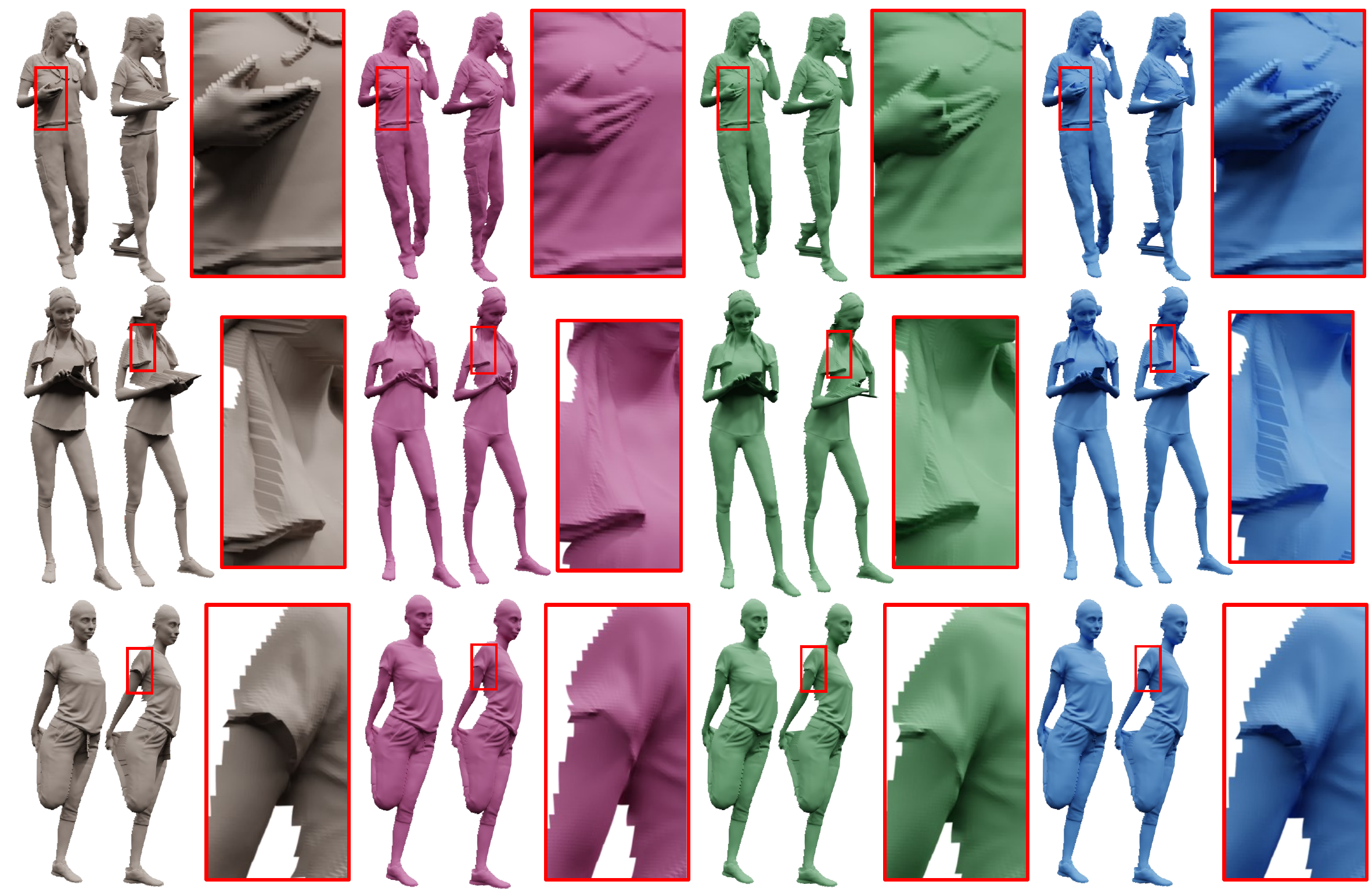}\\
    \hspace{20pt} Ground truth \hspace{70pt} Poisson \cite{horn1986variational} \hspace{80pt} BiNI \cite{cao2022bilateral} \hspace{85pt} Ours \hspace{40pt}
    \caption{Qualitative comparison to previous normal integration methods using the human body dataset.}
    \label{fig:qualitative-human}
\end{figure*}
\section{Experiments}

\subsection{Implementation details}

In the following experiments, we use the hyperparameters $\lambda_{soft}=0.2$, $\lambda_{hard}=1.2$, $N_{max}=5000$, $k=1000$, and $\tau = 10^{-2}$. Our method is implemented using PyTorch with CUDA enabled. Our unoptimized prototype implementation takes approximately 120 seconds to process a normal map of size $512 \times 512$ using a Linux Desktop with Intel i7-10700 CPU and NVIDIA Titan RTX GPU.

\subsection{Comparison with previous methods}

We compare our methods with the state-of-the-art discontinuity-preserving normal integration method, BiNI \cite{cao2022bilateral}, which demonstrated significant improvement over previous discontinuity-preserving normal integration methods \cite{queau_variational_2018} in terms of depth accuracy. We evaluate the accuracy of depth reconstruction using a dataset with known ground-truth 3D meshes for the objects to be reconstructed. We 
use DiLiGenT dataset \cite{shi2016benchmark}, which contains normal maps and ground-truth depths for 10 objects, including sculptures of animals and teapots. We also construct a 3D human body dataset by collecting 27 human body models from RenderPeople \cite{renderpeople}. For the human body dataset, we render the input normal maps and ground-truth depth maps using the ground-truth 3D meshes to build evaluation data. We use a perspective camera with field of view $33.4^\circ$ for rendering.

In \Figs{qualitative-diligent} and \ref{fig:qualitative-human}, we qualitatively compare our method with two methods: Poisson solver \cite{horn1986variational} that does not consider discontinuity in the integration, and the state-of-the-art method BiNI \cite{cao2022bilateral}, using the hyperparameter $k=2$ used in the paper. The Poisson solver introduced global distortions to the reconstruction results, leading to large differences from the ground-truth depth map. BiNI successfully handled large and obvious discontinuities. However, small and challenging discontinuities with small jumps are not handled properly. Our method correctly recovered large discontinuities found by BiNI. In addition, our method successfully recovered small discontinuities that are missed by BiNI. Our method can detect small discontinuities by solving for a more sparse solution based on iterative filtering of explicit gradients on auxiliary edges.

\Tbl{quantitative} shows quantitative evaluation results for the DiLiGenT dataset. Compared to the Poisson solver, our method and BiNI reconstruct the surface accurately; compared to BiNI, our method produces a lower error in most of the cases because jumps for small discontinuities are accurately handled by our sparsity regularization.

\begin{table}[t]
    \caption{Quantitative evaluation results for the DiLiGenT dataset. Mean absolute depth errors from the ground-truth are presented. We compare mainly with BiNI \cite{cao2022bilateral} because this state-of-the-art method demonstrated large improvements over previous discontinuity-preserving methods. Refer to BiNI \cite{cao2022bilateral} for the quantitative evaluation results of other methods.}
    \label{tab:quantitative}
    \centering
    
\resizebox{.995\linewidth}{!}{
    \begin{tabular}{c|c|c|c|c|c|c|c|c|c}
            & {\rotatebox[origin=c]{90}{Bear}}
            & {\rotatebox[origin=c]{90}{Buddha}}
            & {\rotatebox[origin=c]{90}{Cat}}
            & {\rotatebox[origin=c]{90}{Cow}}
            & {\rotatebox[origin=c]{90}{Harvest}}
            & {\rotatebox[origin=c]{90}{Pot1}}
            & {\rotatebox[origin=c]{90}{Pot2}}
            & {\rotatebox[origin=c]{90}{Reading}}
            & {\rotatebox[origin=c]{90}{Goblet}}
            \\
        \hline
         Poisson \cite{horn1986variational}
         & 1.20
         & 3.71
         & 1.60
         & 0.89
         & 10.09
         & 1.51
         & 0.75
         & 6.62
         & 11.64
         \\
         \hline
         BiNI \cite{cao2022bilateral}
         & 0.49
         & 0.76
         & \textbf{0.11}
         & 0.07
         & 3.30
         & 0.63
         & 0.24
         & 0.32
         & \textbf{8.65}
         \\
         \hline
         Ours
         & \textbf{0.45}
         & \textbf{0.67}
         & 0.24
         & \textbf{0.06}
         & \textbf{2.44}
         & \textbf{0.57}
         & \textbf{0.19}
         & \textbf{0.15}
         & 9.02
    \end{tabular}}
    \vspace{-24pt}
\end{table}

\begin{figure}
    \centering
    \includegraphics[width=\linewidth]{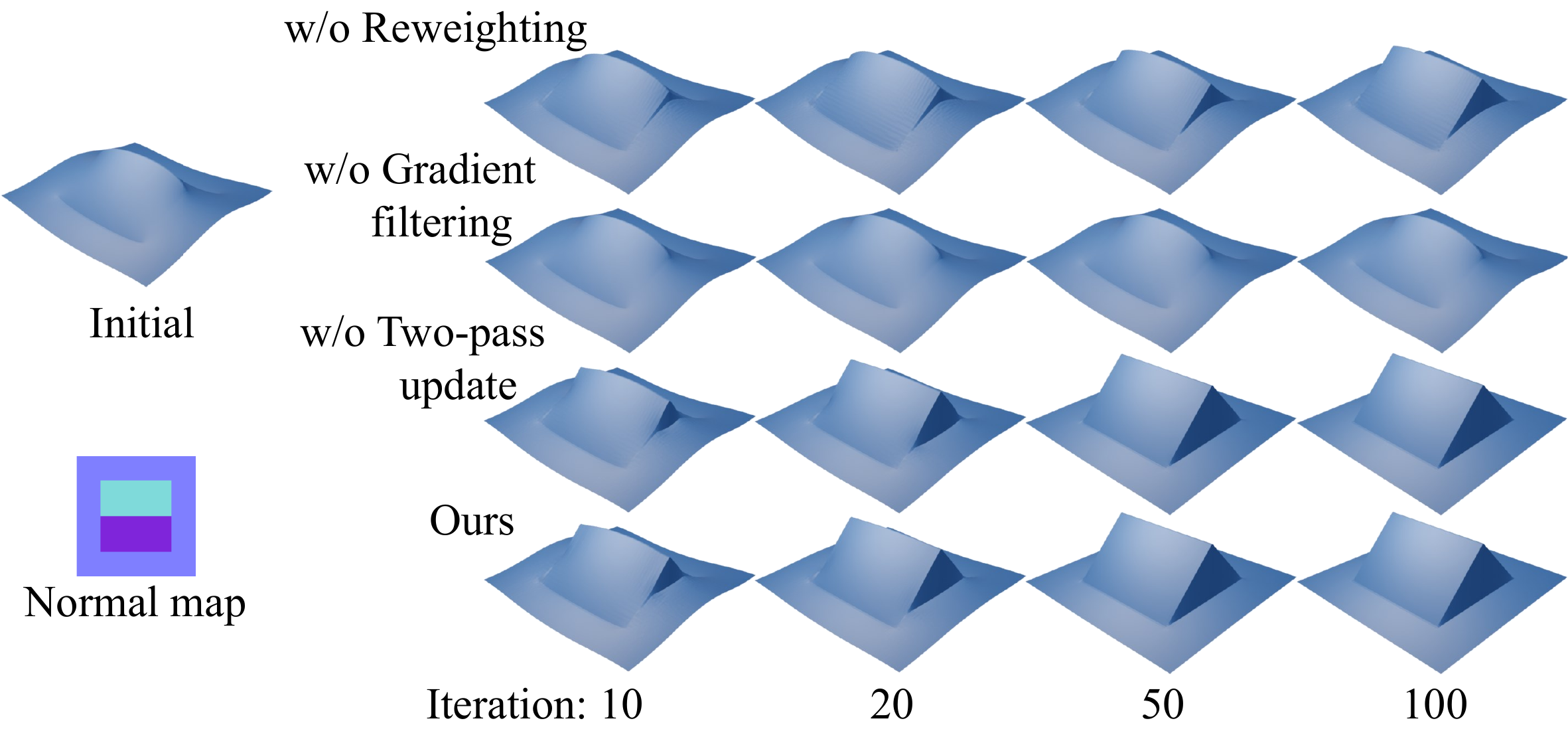}\\
    \includegraphics[width=\linewidth]{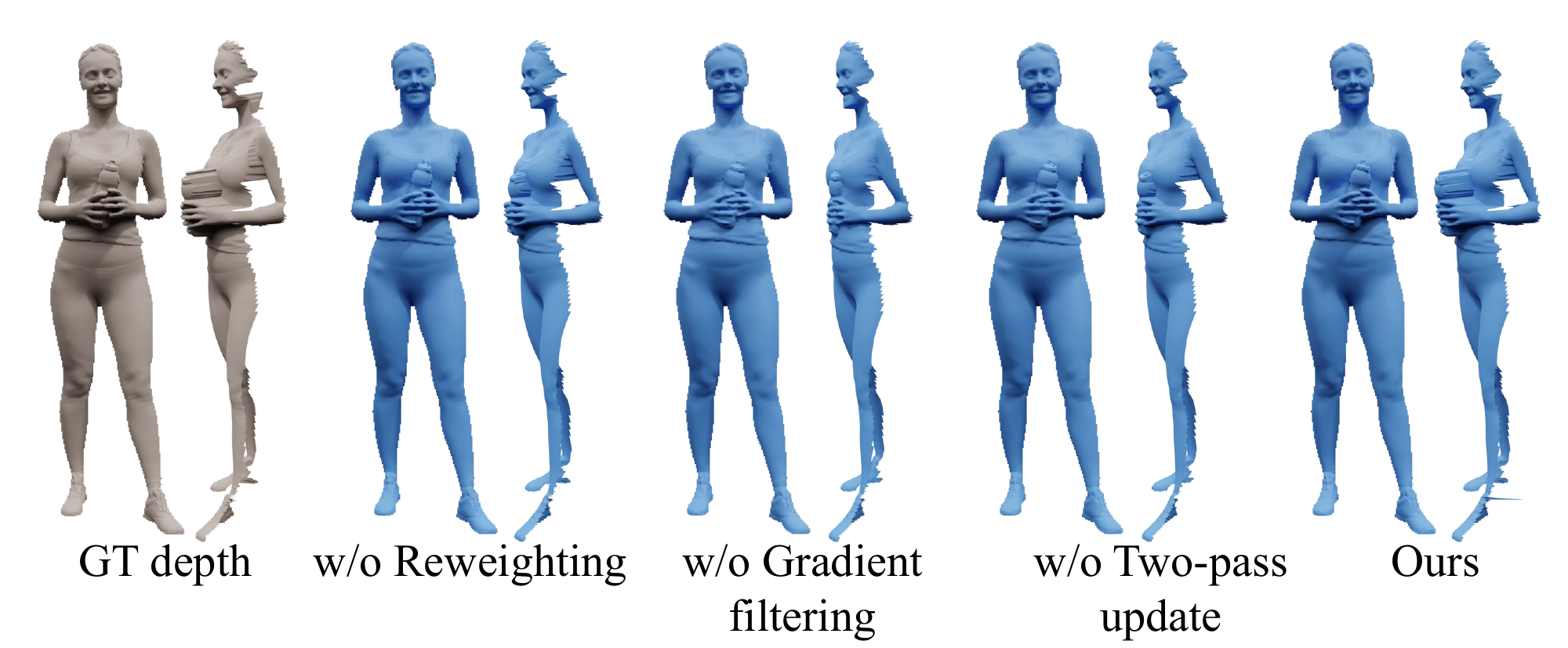}
    \vspace{-12pt}
    \caption{Ablation study of our iterative optimization algorithm. All of the reweighting, the gradient filtering, and the two-pass updates are required to obtain accurate discontinuities.}
    \label{fig:ablation}
    \vspace{-12pt}
\end{figure}

\subsection{Ablation study}
\label{sec:ablation}

\Fig{ablation} shows ablation study results for our iterative optimization. Without the reweighting, the optimization of the discontinuity takes a longer time because the updates for the jump magnitudes become more conservative due to remaining high weights for residuals on auxiliary edges at the discontinuity. Without gradient filtering, the iterative reweighting alone would not effectively create sparse discontinuities. Without two-pass update, the optimization fails to create discontinuities successfully for complex shapes.

\subsection{Building a mesh for the reconstructed surface}
\begin{figure}
    \centering
    \includegraphics[width=\linewidth]{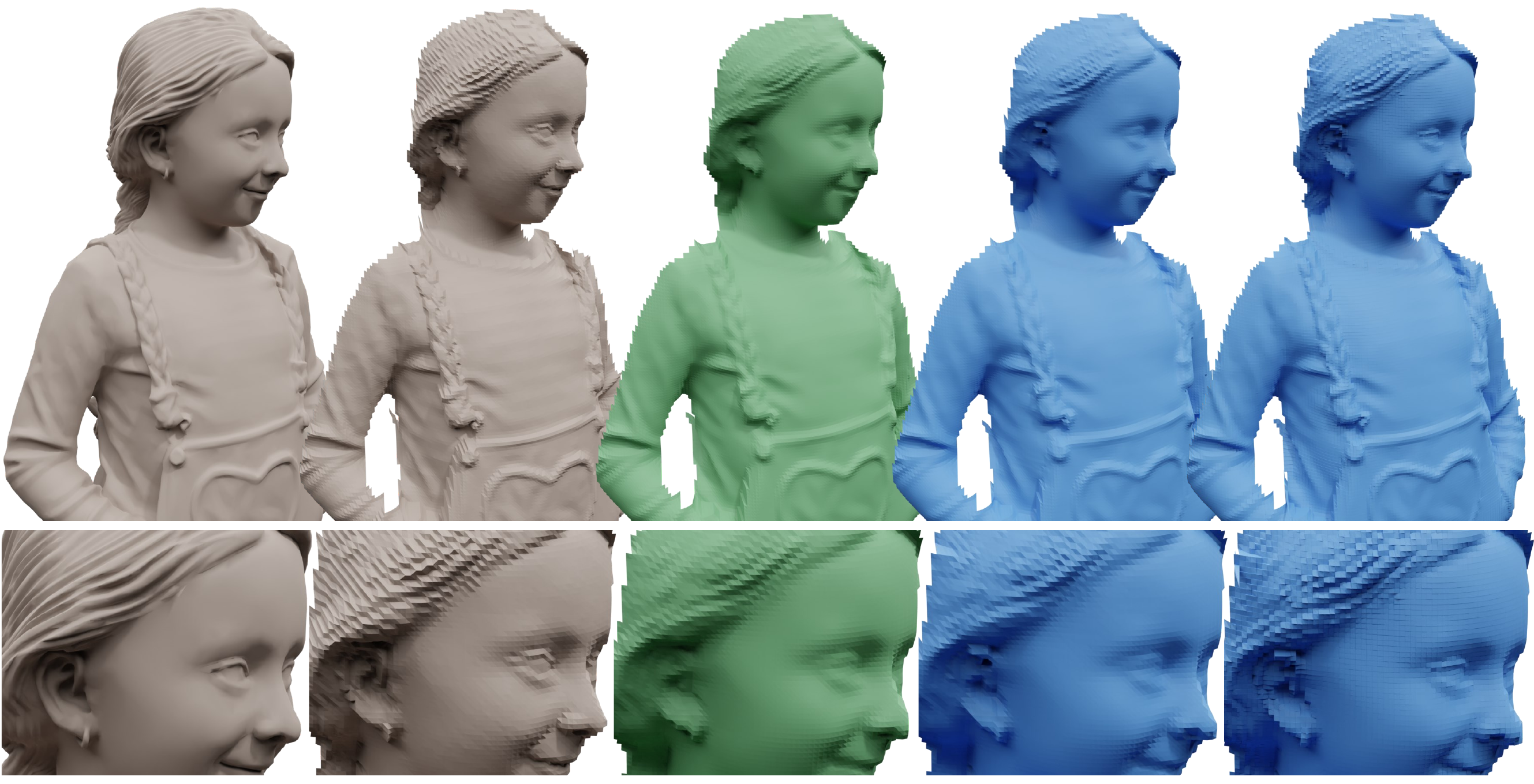}
    \caption{Compared to the usual approach of normal integration using the input image grid, our integration using quadrilaterals on pixels reconstructs a 3D surface with more accurate normals. From left to right: GT mesh that the GT depth map is sampled from, GT depth map, BiNI \cite{cao2022bilateral} result, our result with depth averaging per quadrilateral, and our result without the averaging. The last column shows near-identical shading to the GT mesh and depth.}
    \label{fig:normal-computation}
    \vspace{-6pt}
\end{figure}

After normal integration, we build the final depth map by averaging four depth vertices in each quadrilateral. As a result, the final reconstructed surface is smoothed. We visually find that this smoothing is also observed when solving the normal integration on the original image grid, e.g., as in BiNI \cite{cao2022bilateral}. In contrast, by building a 3D mesh directly using the quadrilaterals in our graph with auxiliary edges, the reconstructed surface can exhibit almost perfect matching to the ground truth surface in terms of appearance. If the output dimension for the normal integration is not restricted to be the input grid, a mesh directly built on our graph would recover details better, as visualized in \Fig{normal-computation}.

\section{Conclusion}

In this paper, we introduced a discontinuity-preserving normal integration method based on a novel discretization scheme of the integration domain. To handle surface discontinuities in normal integration, previous approaches 
implicitly weaken the effects of large gradients across occlusion boundaries. In contrast, we explicitly model the magnitudes of sparse gradients for occlusion boundaries using auxiliary edges. Our approach performs iterative optimization of gradients at auxiliary edges, achieving strong sparsity regularization on the discontinuities. %

\begin{figure}
    \centering
    \includegraphics[width=0.9\linewidth]{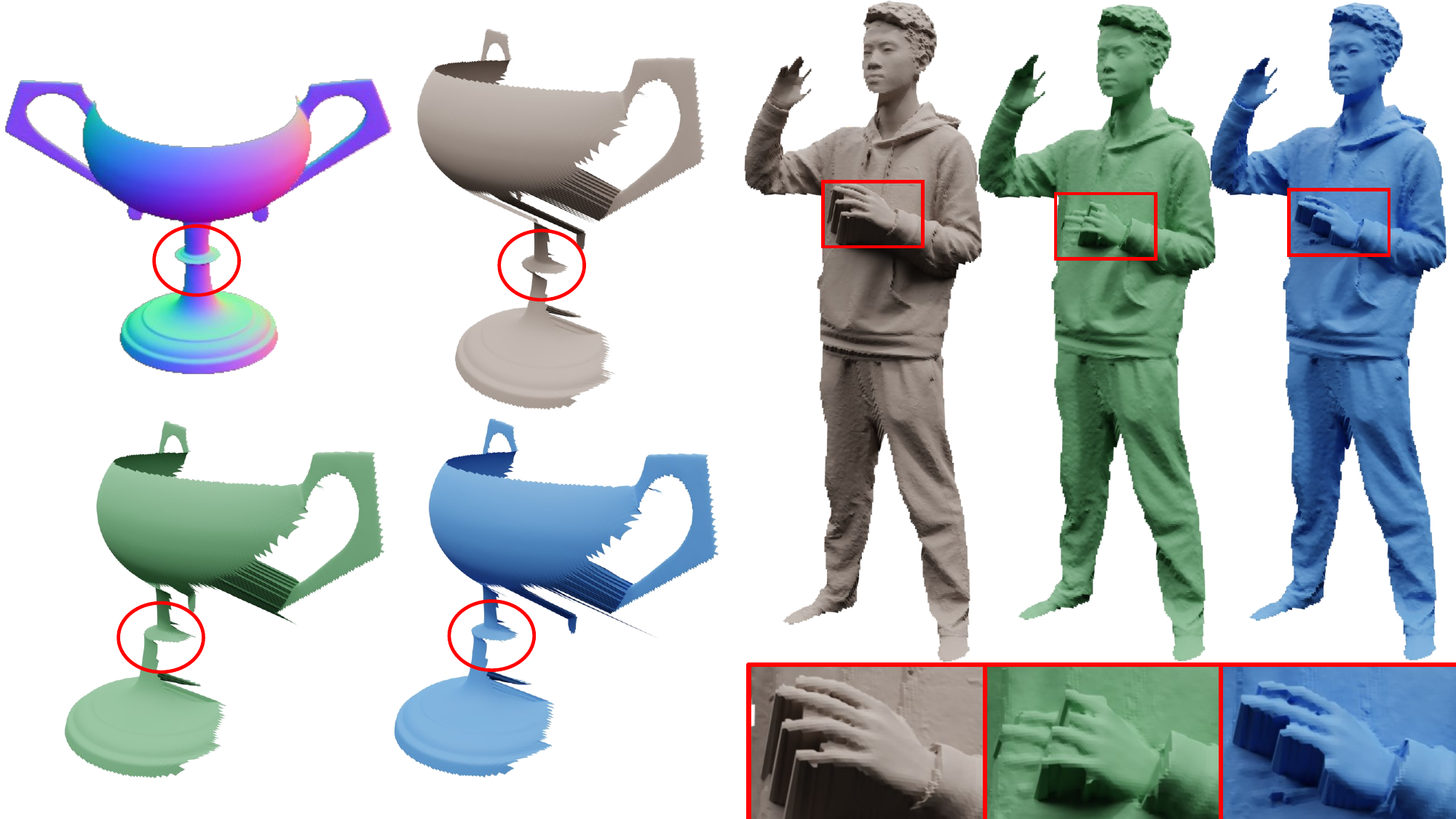}\\
    (a) \hspace{100pt} (b)
    \vspace{-6pt}
    \caption{Failure cases. Brown: ground-truth depth map. Green: BiNI \cite{cao2022bilateral} results. Blue: ours. (a) Due to the global offset or scale ambiguity of the normal integration framework in general, the depth can be erroneous if the discontinuities completely cut the integration domain. (b) Depth reconstruction of noisy human hand scan fails in both BiNI and ours.}
    \label{fig:limitation}
    \vspace{-9pt}
\end{figure}

Other important properties than discontinuity preservation, such as robustness to noise or faster running time discussed in the survey of Quéau~\Etal~\cite{queau_normal_2018}, are left as future work. In addition, a normal map can contain multiple regions completely separated by occlusion boundaries. In this case, due to the ambiguities of normal integration in global translation or scale, large depth errors can occur in such surface islands (\Fig{limitation}a). Handling the depth ambiguity for surface islands can be an interesting research direction.

\vspace*{-0.25cm}
\paragraph{Limitations} 
Our iterative optimization recovers most surface discontinuities correctly, but failure cases exist. For a noisy human scan, reconstruction of fingers is challenging, and both BiNI and our method fail to reconstruct the correct discontinuity (\Fig{limitation}b). Incorporating a depth bias from a data-driven model could be a promising approach for resolving both the aforementioned depth ambiguities and challenging failure cases.

\vspace*{-0.25cm}
\paragraph{Acknowledgements}
This work was supported by the NRF grant (RS-2023-00280400) and IITP grants (AI Graduate School Program, 2019-0-01906; AI Innovation Hub, 2021-0-02068) funded by Korea government (MSIT) and Samsung Electronics.

{
    \small
    \bibliographystyle{ieeenat_fullname}
    \bibliography{main}
}

\end{document}


\maketitle
\blfootnote{$^*$Equal contribution.}

This supplementary document provides additional details for the proposed method and a gallery of results that includes more visual comparisons with other methods
Accompanying supplementary video provides visualization of our iterative optimization.

\section{Normal Maps Captured from a Perspective Camera}
\label{sec:pers}

For simplicity, in the main paper, our formulations for the normal integration and the iterative optimization assume that the normal map has been captured with an orthographic camera. When the normal map is captured with a perspective camera, a correct depth map can be recovered by applying minor modifications to the original formulation.

We employ the formulation for normal integration using the perspective camera in \cite{cao2022bilateral}. Given the normal map captured with a perspective camera of focal length $f$ and image center coordinates $(c_u, c_v)$, the relation between the gradient of the depth map $\mathbf{d}$ and the surface normal $\mathbf{n}$ is
\begin{equation}
    \nabla \Tilde{\mathbf{d}} = \Tilde{\mathbf{g}}
    \label{equ:master}
\end{equation} where
\begin{equation}
    \Tilde{\mathbf{d}} = \log{\mathbf{d}},
    \label{equ:log-depth}
\end{equation}
\begin{equation}
    \Tilde{\mathbf{g}} = \dfrac{\mathbf{n}_{xy}}{\Tilde{n}_z},
\end{equation} and the $z$ component of the normal $n_z$ is adjusted to form $\Tilde{n}_z$ according to image coordinate $(u, v)$:
\begin{equation}
    \Tilde{n}_z = n_x(u - c_u) + n_y(v - c_v) + n_z f.
    \label{equ:normal_z}
\end{equation}
For a more detailed explanation of the derivation of this formulation, refer to \cite{cao2022bilateral}. 

Using this result, the normal integration for the normal map captured from a perspective camera can be solved similarly to the orthogonal case by introducing minor changes. First, the $z$ component of the input normal is modified using \Eq{normal_z}. Then, the log depth map $\Tilde{\mathbf{d}}$ is acquired by performing the normal integration for the orthographic case using the modified normal as the input. Finally, the resulting depth map is obtained with exponential.

\paragraph{Iterative optimization}
Based on the relation between the depth and the gradient in \Eq{master}, our proposed iterative optimization method for the discontinuity and the depth map is modified accordingly. In the perspective case, the data term $E_{data} = E_v + E_a$ is defined as
\begin{equation}
    E_v(\hat{\mathbf{g}}, \Tilde{\mathbf{d}}) = \sum_{e \in \mathcal{E}_v} \lVert \Tilde{n}_z(e)(D_e \Tilde{\mathbf{d}}  - \Tilde{\mathbf{g}}(e)) \rVert^2,
\end{equation}
\begin{equation}
    E_a(\mathbf{g}', \Tilde{\mathbf{d}}) = \sum_{e \in \mathcal{E}_a} \lVert \Tilde{n}_z'(e)(D_e \Tilde{\mathbf{d}}  - \mathbf{g}'(e)) \rVert^2,
\end{equation}
and subsequently, the energy $E_{disc}$ becomes
\begin{equation}
    E_{disc}(\mathbf{g}', \Tilde{\mathbf{d}}, \mathbf{w}) = \sum_{e \in \mathcal{E}_a} w_{e}\lVert \Tilde{n}_z'(e)(D_e \Tilde{\mathbf{d}}  - \mathbf{g}'(e))\rVert^2,
\end{equation}
where the value $\Tilde{n}_z'$ is simply a constant $\Tilde{n}_z' = f$. Note that this value is the result of \Eq{normal_z} when the normal is $(n_x, n_y, n_z) = (0, 0, 1)$. %

\section{Results for Various Inputs}
\begin{figure}
    \centering
    \includegraphics[width=0.98\linewidth]{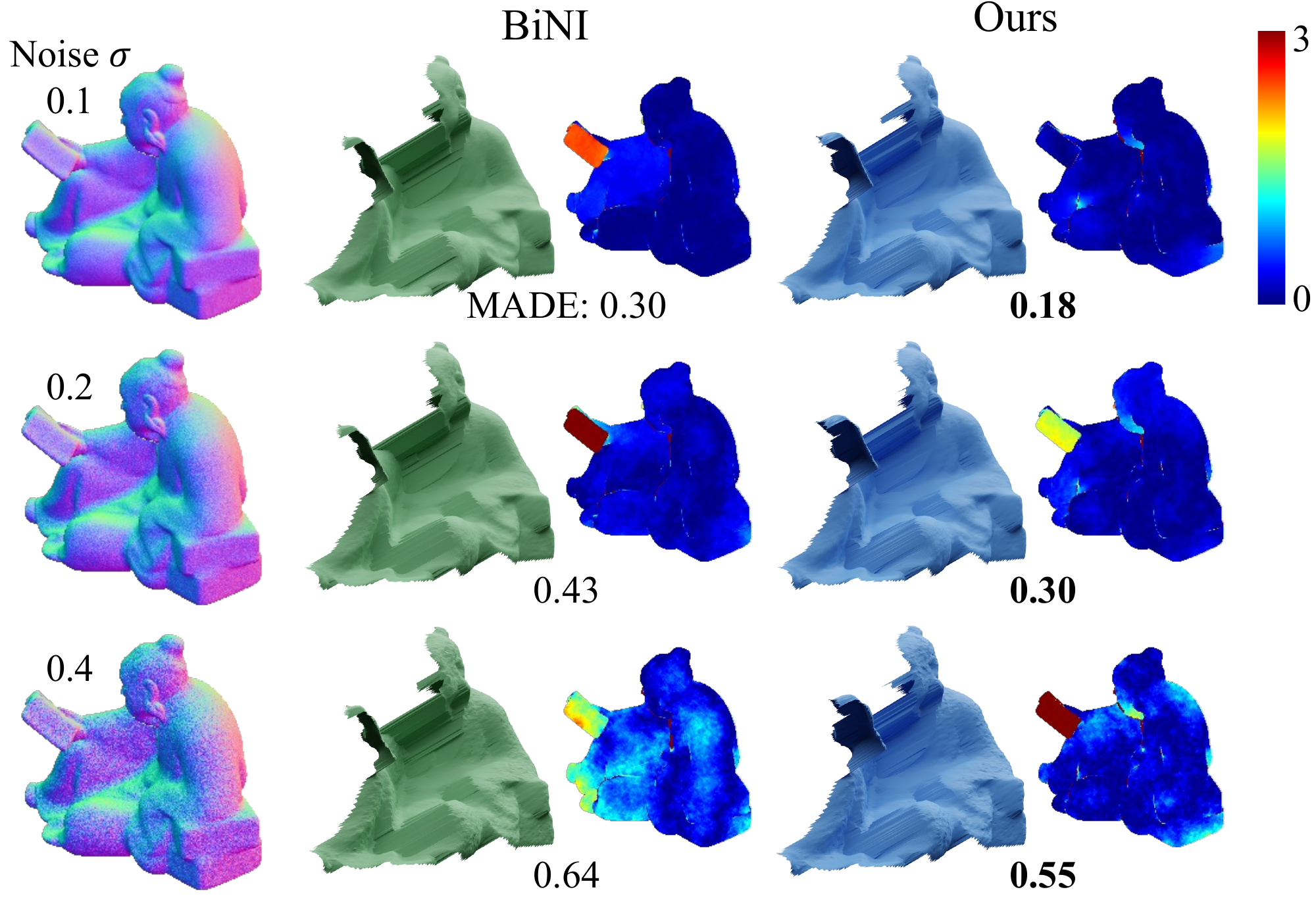}
    \caption{
    Normal integration when zero-mean Gaussian noise with different standard deviations $\sigma$ is added to gradients. Our method produces plausible visual results in all three noise levels. Our method also shows better accuracy than BiNI \cite{cao2022bilateral} in terms of mean absolute depth error (MADE) to the GT depth map.
    }
    \label{fig:noise}
\end{figure}

\begin{figure}
    \centering
    \includegraphics[width=0.98\linewidth]{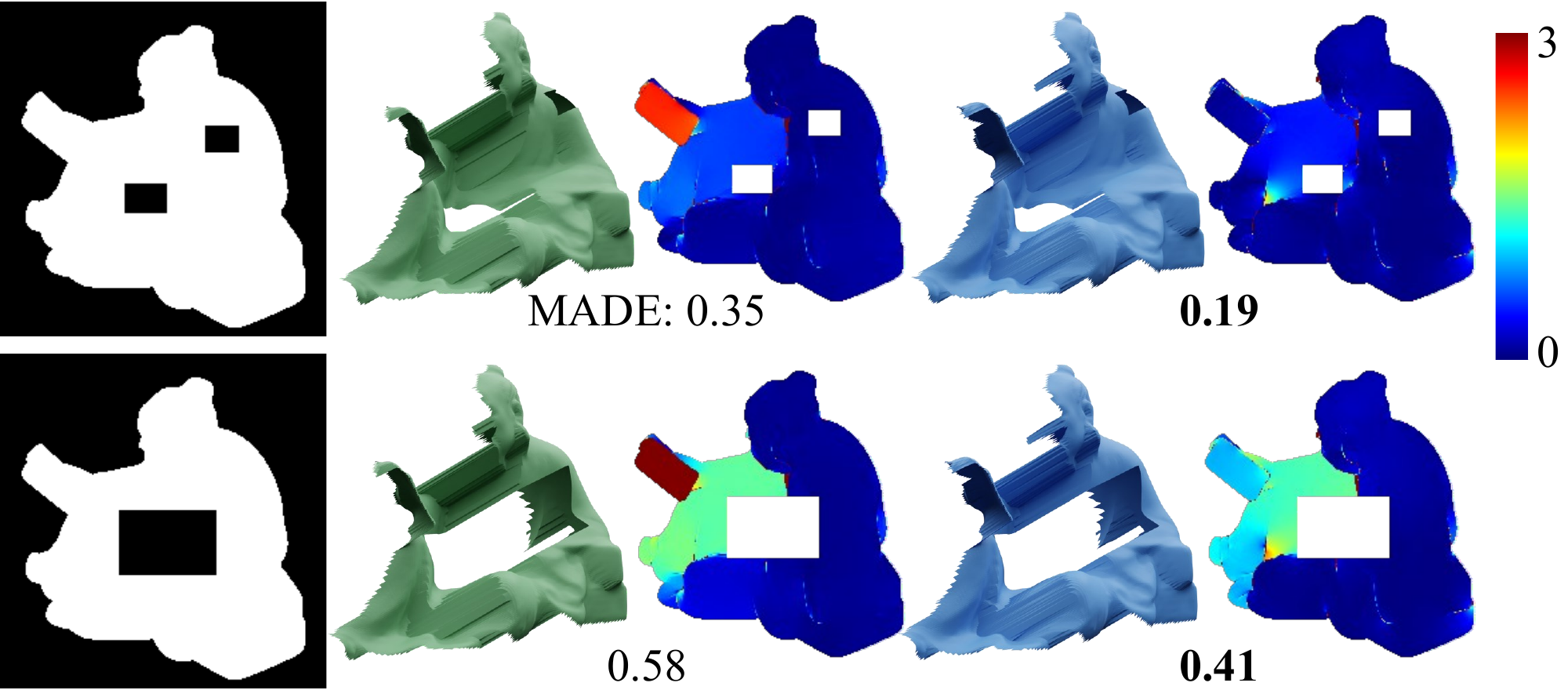}
    \caption{Normal integration results for surfaces with holes. Our method produces better results than BiNI \cite{cao2022bilateral} in terms of mean absolute depth error (MADE).
    }
    \label{fig:hole}
\end{figure}

\paragraph{Inputs with noise and hole}
\Fig{noise} presents results for normal maps with noise. Our method is robust to noise because our method uses combination of explicit gradient filtering and least-squares method; when the gradient filtering produces faulty gradient edit $\mathbf{g}'$ due to noise, the least-squares in the depth solving step can heal the damage. \Fig{hole} presents results for normal maps with holes. Our method robustly handles normal maps containing small holes (top row) and a large hole (bottom row).

\begin{figure}
    \centering
    \includegraphics[width=0.98\linewidth]{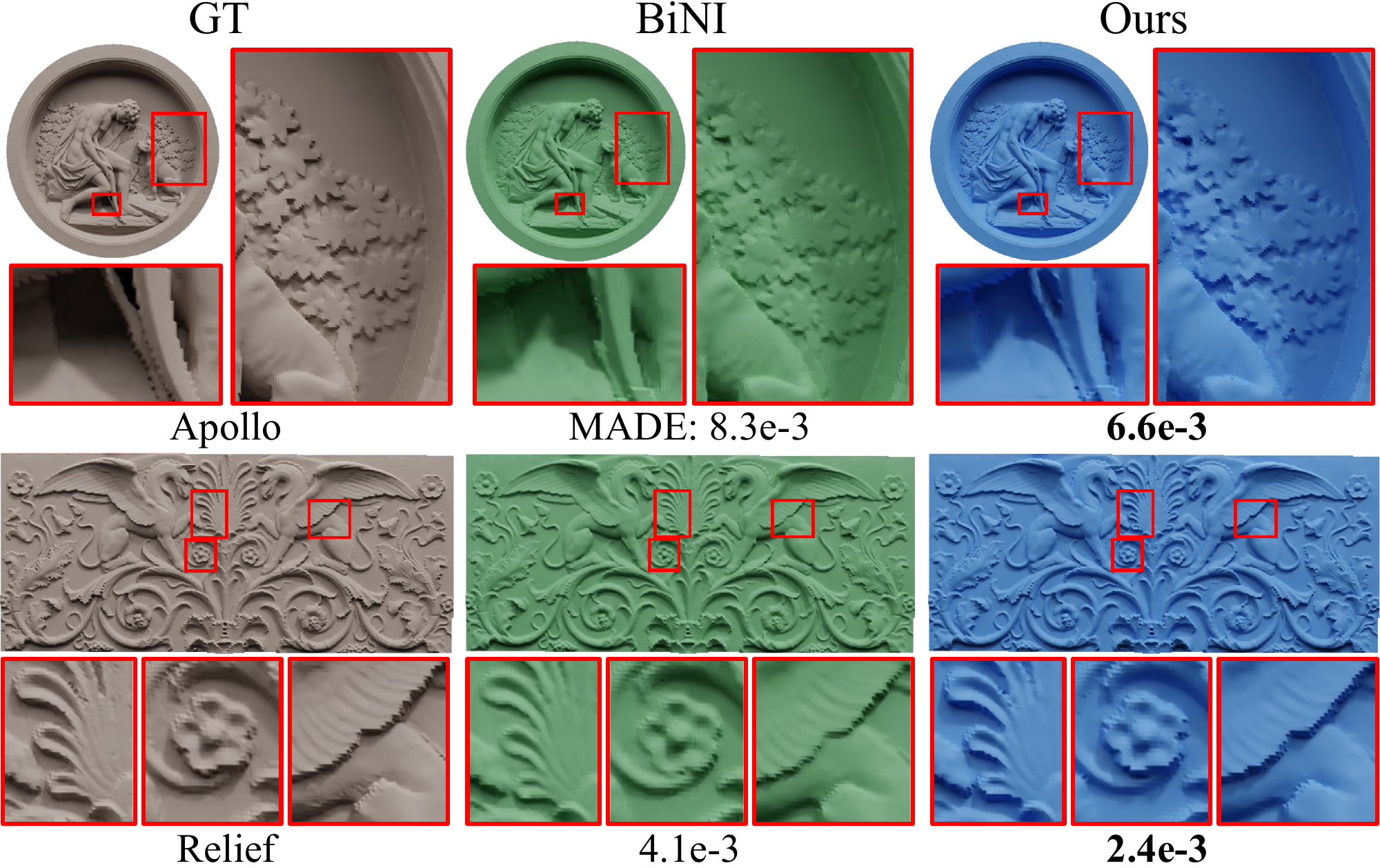}
    \caption{
    Normal integration results for surfaces with details\protect\footnotemark[1] \protect\footnotemark[2]. Our method produces better visual results capturing subtle GT details as can be observed with the appearance of shadows. Our method also produces better results than BiNI \cite{cao2022bilateral} in terms of mean absolute depth error (MADE). 
    }
    \label{fig:details}
\end{figure}

\paragraph{Inputs with surface details}
\Fig{details} presents results for two bas-relief normal maps. Bas-relief contains many subtle details that introduce discontinuities. Our method handles such subtle details correctly.

\begin{figure}
    \centering
    \includegraphics[width=0.98\linewidth]{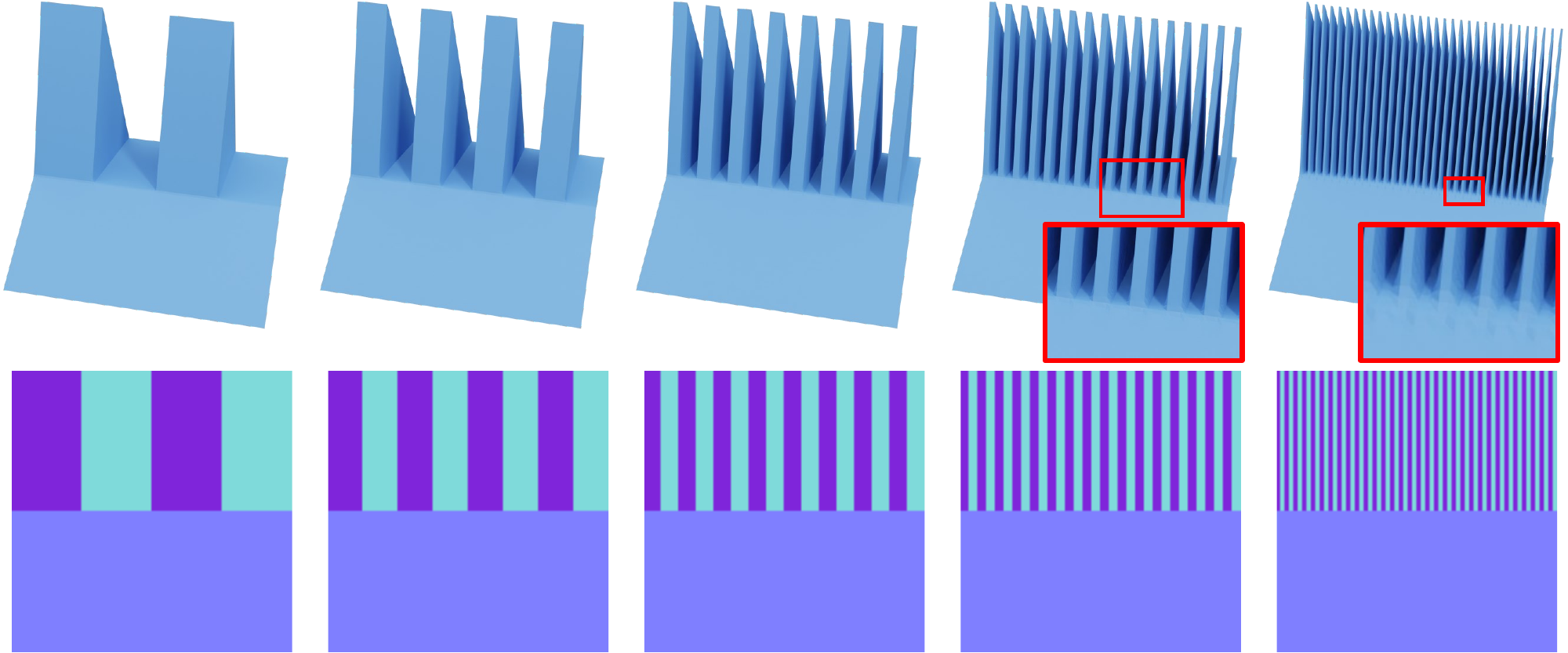}
    \caption{
    Our results for comb-like normal maps. The input normal map for the third column is the same as the input used in BiNI's Fig.\ 4 \cite{cao2022bilateral}. Our method produces correct results for variations of comb-like normal maps with different frequencies.}
    \label{fig:inputs}
\end{figure}

\paragraph{Objects with different surface frequencies}
In Table 1 of the main paper, for instances with mostly smooth surfaces ({\em Bear}, {\em Cat}, {\em Cow}), our results have similar errors to BiNI's. For instances with mid-scale details ({\em Buddah}, {\em Harvest}, {\em Reading}), our results have significantly lower errors than BiNI's. We show examples and evaluate the errors for surfaces with fine details in \Fig{details}.
We visualise results for different frequencies of teeth in~\Fig{inputs}. We obtained better or similar results to BiNI \cite{cao2022bilateral} in all cases. The comb-like normal maps contain sparse discontinuities only at the boundaries of the teeth.

\section{Experimental Analysis of Hyperparameters}

All examples in our main paper and the supplementary use the same hyperparameters. \Tbl{quan_hyper} shows quantitative evaluation results for the DiLiGenT dataset with different hyperparameters. Parameters except $\lambda_{soft}$ and $\lambda_{hard}$ can stay fixed as they do not change results notably.

\footnotetext[1]{"Relief", CC BY 4.0. \href{https://sketchfab.com/3d-models/relief-47f53fd51c7042c59b42ffc8b01621ad}{https://sketchfab.com/3d-models/relief-47f53fd51c7042c59b42ffc8b01621ad}}
\footnotetext[2]{"Apollo", CC BY 4.0. \href{https://sketchfab.com/3d-models/apollo-941a94bfba164dc6a9f682d4820e0dae}{https://sketchfab.com/3d-models/apollo-941a94bfba164dc6a9f682d4820e0dae}}

\begin{table}
    \caption{Quantitative evaluation results for the DiLiGenT dataset with different hyperparameters. For all examples in our main paper and the supplementary, we use the hyperparameters $k=1000, \lambda_{soft}=0.2, \lambda_{hard}=1.2$ (Ours, first row).}
    \label{tab:quan_hyper}
    \centering
    
\resizebox{.995\linewidth}{!}{
    \begin{tabular}{c|c|c|c|c|c|c|c|c|c}
            & {\rotatebox[origin=c]{90}{Bear}}
            & {\rotatebox[origin=c]{90}{Buddha}}
            & {\rotatebox[origin=c]{90}{Cat}}
            & {\rotatebox[origin=c]{90}{Cow}}
            & {\rotatebox[origin=c]{90}{Harvest}}
            & {\rotatebox[origin=c]{90}{Pot1}}
            & {\rotatebox[origin=c]{90}{Pot2}}
            & {\rotatebox[origin=c]{90}{Reading}}
            & {\rotatebox[origin=c]{90}{Goblet}}
            \\
        \hline
         Ours&&\cellcolor{yellow!20}{}&&\cellcolor{yellow!100}{}&\cellcolor{yellow!100}{}&\cellcolor{yellow!100}{}&\cellcolor{yellow!20}{}&\cellcolor{yellow!100}{}&\\($k, \lambda_{soft}, \lambda_{hard}$)
         & \multirow{-2}{*}{0.45}
         & \multirow{-2}{*}{\cellcolor{yellow!20}{0.67}}
         & \multirow{-2}{*}{0.24}
         & \multirow{-2}{*}{\cellcolor{yellow!100}{\textbf{0.06}}}
         & \multirow{-2}{*}{\cellcolor{yellow!100}{\textbf{2.44}}}
         & \multirow{-2}{*}{\cellcolor{yellow!100}{\textbf{0.57}}}
         & \multirow{-2}{*}{\cellcolor{yellow!20}{0.19}}
         & \multirow{-2}{*}{\cellcolor{yellow!100}{\textbf{0.15}}}
         & \multirow{-2}{*}{9.02}\\
        \hline
         (100, 0.2, 1.2)
         & 0.49
         & \cellcolor{yellow!100}{\textbf{0.41}}
         & 0.20
         & 0.21
         & \cellcolor{yellow!20}{2.94}
         & 0.60
         & \cellcolor{yellow!100}{\textbf{0.17}}
         & 0.42
         & 9.15 \\
        \hline
         (2000, 0.2, 1.2)
         & 0.48
         & 0.78
         & 0.23
         & \cellcolor{yellow!20}{0.10}
         & 2.98
         & 0.62
         & 0.20
         & 0.16
         & 9.00 \\
        \hline
         (100, 0.1, 1.1)
         & 0.36
         & 1.64
         & \cellcolor{yellow!100}\textbf{0.14}
         & 0.13
         & 4.38
         & \cellcolor{yellow!20}{0.58}
         & 0.20
         & 0.40
         & 9.04 \\
        \hline
         (1000, 0.1, 1.1)
         & \cellcolor{yellow!20}{0.30}
         & 2.40
         & 0.16
         & 0.13
         & 4.93
         & 0.70
         & 0.24
         & 0.52
         & \cellcolor{yellow!20}{7.95} \\
        \hline
         (2000, 0.1, 1.1)
         & \cellcolor{yellow!100}{\textbf{0.14}}
         & 2.44
         & \cellcolor{yellow!20}{0.15}
         & 0.15
         & 5.21
         & 0.81
         & 0.25
         & 0.65
         & \cellcolor{yellow!100}{\textbf{7.90}} \\
    \end{tabular}}
\end{table}

\section{Additional Technical Details}

\paragraph{GT depth from GT mesh in the main paper's Fig. 8}
GT mesh is the original 3D mesh from which the input normal map is rendered. GT depth is the visualization of the depth map with the same dimension as the rendered normal map, using each pixel as the vertex.

\paragraph{Time consumption and convergence} 
For the DiLiGeNT dataset, BiNI took 10 seconds on average to converge, and additional iterations do not notably enhance the accuracy. Our method takes 6 seconds on average to reach the same level of depth error as BiNI, and produces more accurate results with additional iterations. Using a fixed number of iterations $N_{max}=1300$, which takes approximately 50 seconds, our method achieves higher accuracy than BiNI in most cases.

\section{Gallery of Results}
\label{sec:gallery}

\begin{figure*}
    \centering
    \includegraphics[width=1.00\textwidth]{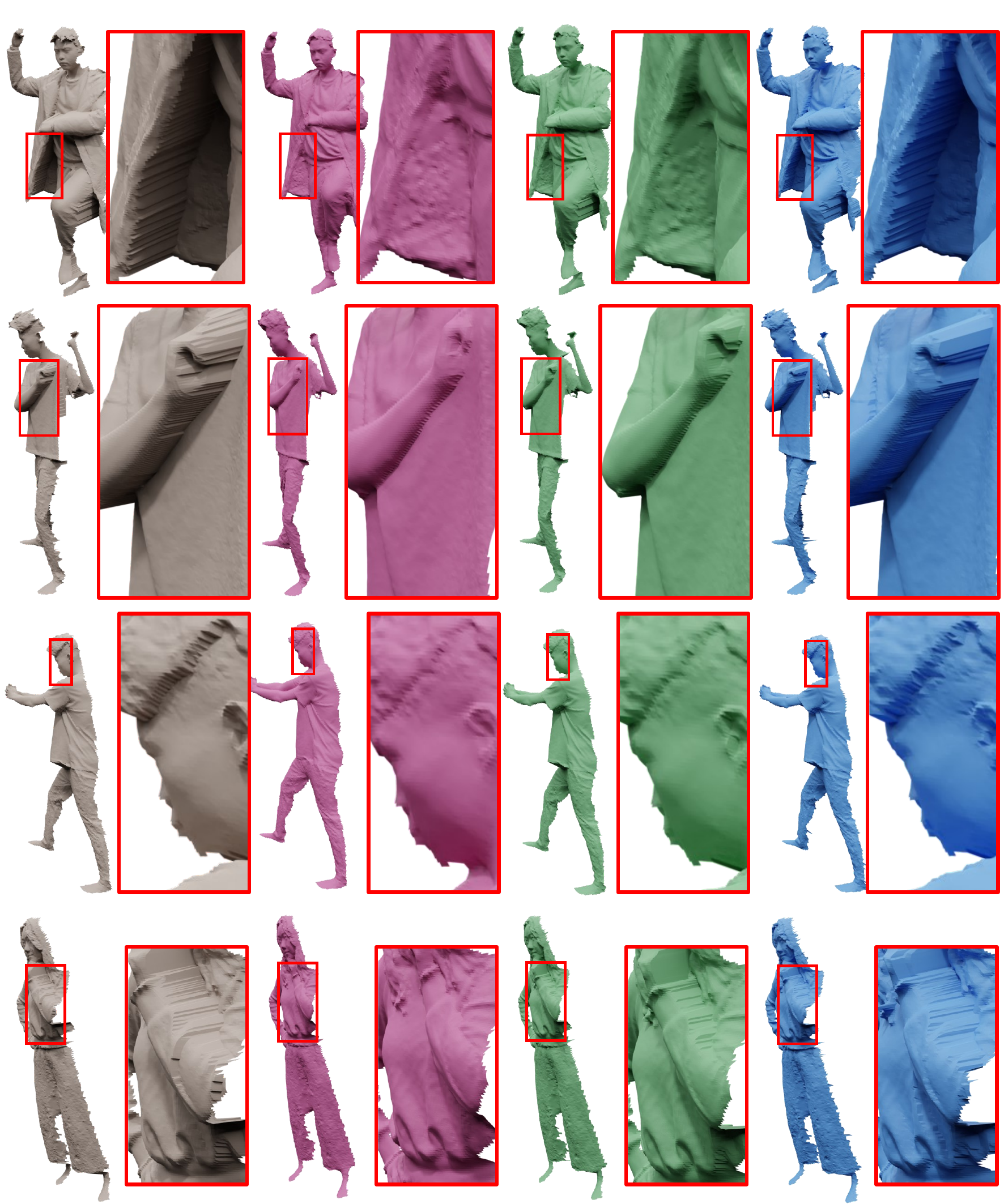}\\
    \hspace{20pt} Ground truth \hspace{70pt} Poisson \cite{horn1986variational} \hspace{80pt} BiNI \cite{cao2022bilateral} \hspace{85pt} Ours \hspace{40pt}\\
    \caption{Qualitative comparison to previous normal integration methods using a human body dataset.}
    \label{fig:gallery1}
\end{figure*}

\begin{figure*}
    \includegraphics[width=1.00\textwidth]{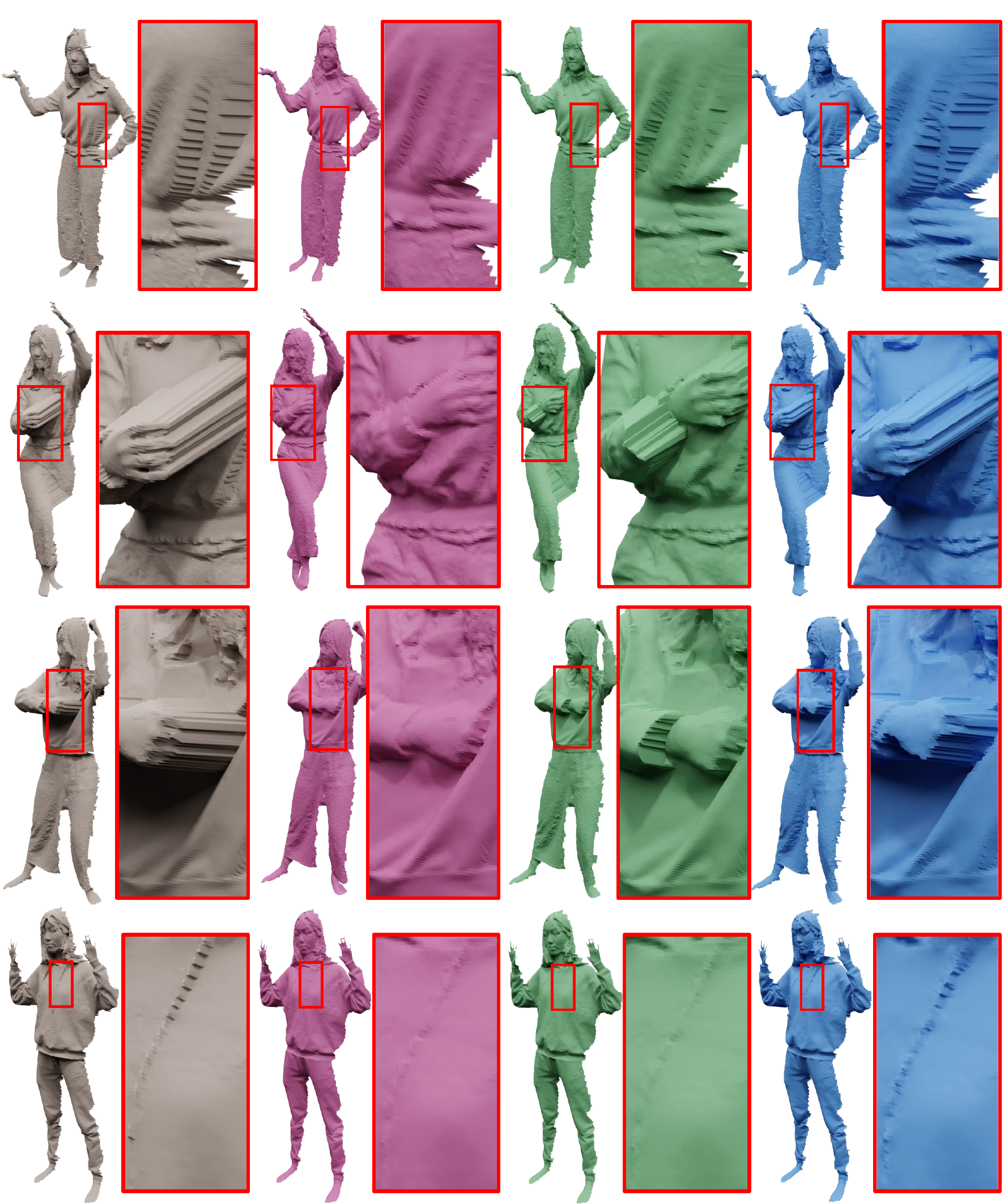}\\
    \hspace{20pt} Ground truth \hspace{70pt} Poisson \cite{horn1986variational} \hspace{80pt} BiNI \cite{cao2022bilateral} \hspace{85pt} Ours \hspace{40pt}\\
    \caption{Qualitative comparison to previous normal integration methods using a human body dataset.}
    \label{fig:gallery2}
\end{figure*}

\begin{figure*}
    \includegraphics[width=1.00\textwidth]{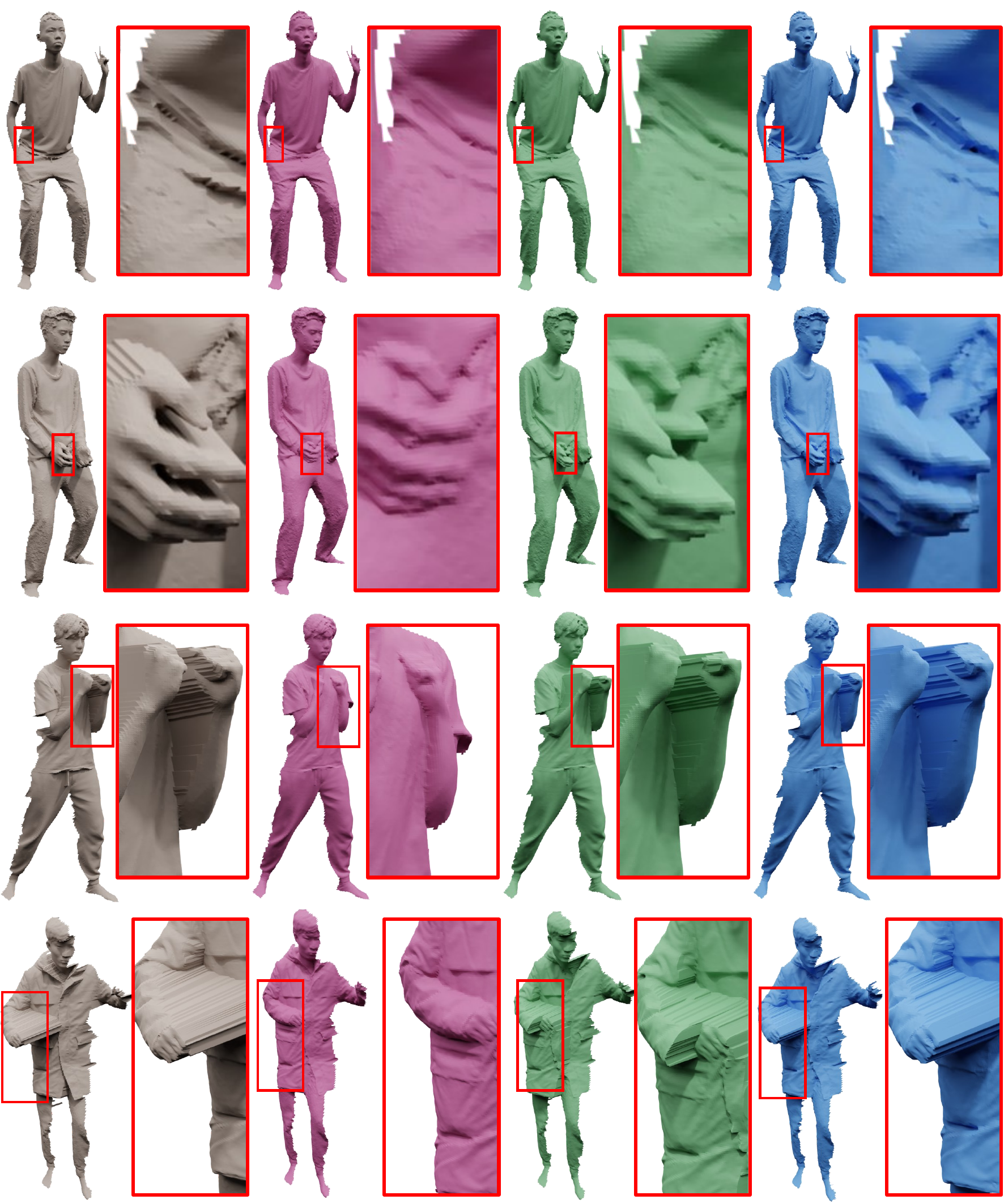}\\
    \hspace{20pt} Ground truth \hspace{70pt} Poisson \cite{horn1986variational} \hspace{80pt} BiNI \cite{cao2022bilateral} \hspace{85pt} Ours \hspace{40pt}
    \caption{Qualitative comparison to previous normal integration methods using a human body dataset.}
    \label{fig:gallery3}
\end{figure*}

In Figs.~\ref{fig:gallery1}, \ref{fig:gallery2}, and \ref{fig:gallery3}, we provide additional visual comparisons to the Poisson method \cite{horn1986variational} and BiNI \cite{cao2022bilateral} using normal maps in THuman2.0~\cite{tao2021function4d} dataset. For BiNI, we use the hyperparameter $K=2$ used in the paper. Due to our explicit representation of the jumps across discontinuities, our method recovers sparse discontinuities accurately even in small and subtle jumps, e.g., across wrinkles of clothes and hair.

{
    \small
    \bibliographystyle{ieeenat_fullname}
    \bibliography{main}
}